\pdfoutput=1
\documentclass[10pt,twocolumn]{article}
\usepackage{multirow}

\usepackage[letterpaper,margin=0.75in,columnsep=0.25in]{geometry}

\usepackage[utf8]{inputenc}
\usepackage[T1]{fontenc}
\usepackage{amsmath,amssymb,amsfonts}
\usepackage{graphicx}
\usepackage{subcaption}
\usepackage{booktabs}
\usepackage{microtype}
\usepackage{enumitem}
\newcommand{\Hecto}{Hecto}
\usepackage{hyperref}
\usepackage{url}
\usepackage{titlesec}
\usepackage{adjustbox}
\usepackage{float}
\usepackage{adjustbox}  

\titlespacing*{\section}{0pt}{1.5ex plus 1ex minus .2ex}{0.8ex plus .2ex}
\titlespacing*{\subsection}{0pt}{1.2ex plus .8ex minus .2ex}{0.6ex plus .2ex}

\begin{document}

\twocolumn[{
  \begin{center}
    \rule{\linewidth}{1pt}\par
    \vspace{0.4cm}
{\LARGE\bfseries Hecto: Modular Sparse Experts for Adaptive and Interpretable Reasoning}

    \vspace{0.3cm}
    \rule{\linewidth}{1pt}\par
    \vspace{0.6cm}
    {\large
      Sanskar Pandey\textsuperscript{*},\;
      Ruhaan Chopra\textsuperscript{*},\;
      Saad Murtaza Bhat,\;
      Ark Abhyudaya\;
     
    }\par
    \vspace{0.3cm}
    {\small
      \textsuperscript{*}Equal contribution. Correspondence to:\\
      \texttt{pandeysanskar854@gmail.com}, 
      \texttt{ruhaanchopra2005@gmail.com},\\
      \texttt{saadmbhat@gmail.com},
      \texttt{arkabhyudaya@gmail.com}\par
      
    }\par
    \vspace{0.6cm}
  \end{center}
}]

\begin{abstract}

Mixture-of-Experts (MoE) models enable conditional computation by routing inputs to specialized experts, but these experts rely on identical inductive biases, thus limiting representational diversity. This static computation pathway is inefficient for inputs that require different types of reasoning and limits specialization and interpretability. We propose \textbf{Hecto}, a lightweight MoE architecture that leverages architectural heterogeneity by combining a GRU expert for temporal reasoning and an FFNN expert for static abstraction under a sparse Top-1 gating mechanism. Evaluated on three reasoning benchmarks (AG News, SST-2, HotpotQA) and a regression task (STS-B), Hecto matches or closely trails homogeneous baselines in performance despite receiving isolated input representations, while achieving clear expert specialization, with each expert aligning to distinct reasoning types (temporal vs static). At larger batch sizes, Hecto exhibits improved performance, benefiting from relaxed computational constraints that allow its heterogeneous architecture to optimize more effectively. Ablation results isolate architectural diversity as the source of Hecto’s stability and interpretability across diverse reasoning tasks. Overall, Hecto establishes itself as a new benchmark for conditional computation, offering a principled framework for specialized reasoning in low-resource regimes with its model strength derived from principled specialization.

\end{abstract}


\section{Introduction}

Standard neural architectures employ a fixed computational pathway for all inputs, regardless of complexity. This inflexibility often leads to under processing of examples that require multi-step or temporal reasoning, and over processing of trivial or static inputs. Such inefficiencies are prominent in tasks with multiple distinct reasoning demands - e.g., factual classification vs. multihop inference.

Mixture-of-Experts (MoE) models address this limitation via \emph{conditional computation}, selectively activating only a subset of modules per input~\cite{shazeer2017outrageously,fedus2022switch}. However, most existing MoEs rely on \emph{homogeneous} expert pools, typically shallow feedforward networks, limiting both functional specialization and interpretability. As a result, these models often struggle to flexibly align the computational structure with input semantics and task requirements.

The two primary categories of representational inference are static reasoning (pattern recognition in fixed representations) and temporal reasoning (contextual inference over sequences), and these correspond to different semantic structures where the former uses fixed embeddings or summary states, and the latter relies on evolving token sequences. These reasoning paradigms are effectively modeled by FFNNs and GRUs. The handling of static and temporal reasoning is necessary for approaching tasks with heterogeneous reasoning demands, where inputs may vary in complexity and structure. The combination of an FFNN expert with a GRU expert is a measured preference aimed at supporting both static and sequential/temporal reasoning.

In this work, we introduce \textbf{Hecto} - a lightweight, modular, and interpretable MoE architecture that combines structural heterogeneity with sparse routing. Built upon a DistilBERT encoder (frozen or fine-tuned), Hecto is an integration of two complementary experts: a GRU for temporal modeling and an FFNN for static abstraction with isolated input representations: a projected sequence for the GRU and [CLS] vector for the FFNN. A Top-1 gating network, regularized via entropy and load-balancing penalties, routes each input to the expert best suited to its computational profile.

\vspace{0.5em}
\noindent Our contributions are as follows.
\begin{enumerate}[leftmargin=*]
\item \textbf{Heterogeneous Expert Pairing:} We introduce a novel MoE configuration comprising a GRU for sequential reasoning and an FFNN for static classification, capturing distinct cognitive modes within a unified framework.

\item \textbf{Interpretable Sparse Gating:} We design a Top-1 gating mechanism with entropy and diversity regularization, favoring confident, balanced expert usage and interpretable routing patterns.

\item \textbf{Robust Modularization:} Hecto supports frozen and fine-tuned encoder regimes and dynamically adapts its execution based on input complexity and depth, enabling parameter-efficient deployment.

\item \textbf{Empirical Validation:} Across various benchmarks such as AG News, SST-2, and HotpotQA, Hecto achieves competitive or superior performance compared to homogeneous MoEs, reducing inference overhead and improving routing interpretability.

\item \textbf{Generalization under Constraints:} Even with a frozen encoder (Appendix~\ref{app:frozen}), Hecto maintains high accuracy (87.76\%) and functionally aligned expert selection (7.5\% FFNN / 92.5\% GRU).
\end{enumerate}

\vspace{0.5em}
\noindent In general, Hecto offers a principled blueprint for modular neural computation, allowing models to adaptively reason not only \emph{what} to compute, but also \emph{how} and \emph{when} to compute.
\label{sec:introduction}
\section{Related Work}

\subsection{Conditional Computation and Sparse MoEs}

Mixture-of-Experts (MoE) models support conditional computation by routing inputs to a sparse subset of experts, enabling efficient processing. Switch Transformer~\cite{fedus2022switch} and GShard~\cite{lepikhin2020gshard} pioneered large scale sparse routing with homogeneous FFNs but lacked expert diversity and interpretability. Outrageously Large Networks~\cite{shazeer2017outrageously}
 demonstrated the scalability of sparse MoEs, while DeepSeekMoE~\cite{deepseekmoe2024} and DynaMoE~\cite{dynamo2022} advanced adaptive computation without structural heterogeneity. Expert Choice Routing~\cite{zhou2022expertchoice}
 and SparseMixer~\cite{sparsemixer2023} offer alternative routing methods yet often retain uniform architectures. Hecto employs a structurally heterogeneous design, GRU and FFNN, combined with entropy and diversity regularization for interpretable, task-aligned routing, contrasting recent prompt-tuned specializations like Spectra~\cite{spectra2024}.

\subsection{Heterogeneity and Modular Design}

Although MoE heterogeneity can enhance expressiveness, many models use static or homogeneous configurations. MoFE~\cite{mofe2024} and ModuleFormer~\cite{moduleformer2023} mix architecture types in fixed layouts. AutoMoE~\cite{automoe2022} and HetuMoE~\cite{hetumoe2022} add hardware awareness but lack explicit role divergence. PromptMoE~\cite{promptmoe2024} and Modular Prompting~\cite{modularprompting2024} explore input-driven expert activation but don't achieve architectural variety. SparseMixer~\cite{sparsemixer2023} introduces sparsity in routing but with similar expert types. In contrast, Hecto dynamically routes inputs across heterogeneous experts with explicit structural specialization. MoE-X~\cite{moex2024} also addresses interpretability with intrinsic design, supporting our choice of built-in gating visualization. Together, these works emphasize the need for modular, adaptable, and interpretable MoE design.

\subsection{Expert Specialization and Role Differentiation}

MoEs often rely on emergent behavior to achieve expert specialization. Models like DeepSpeed-MoE~\cite{rajbhandari2022deepspeed}, ExpertRAG~\cite{expertrag2024}, and DeRS~\cite{ders2024} optimize routing and reuse, but experts remain structurally uniform. DynaMoE~\cite{dynamo2022} introduces task-aware routing, while Spectra~\cite{spectra2024} explores specialization via prompt tuning. MoE-X~\cite{moex2024} targets interpretability through sparsity and functional separation, aligning with Hecto’s explicit role divergence between GRU and FFNN experts. Unlike many MoEs that treat experts as interchangeable, Hecto enforces functional differentiation through entropy and diversity losses, yielding interpretable and specialized processing pathways. To assess specialization across reasoning types, we include an \textbf{SST-2} dataset~\cite{socher2013recursive} in our evaluation, a benchmark that focuses on binary sentiment classification of movie reviews and tests a model’s ability to capture semantic nuance.

\subsection{Interpretability in MoE Models}

Interpretability remains a challenge in MoEs, where hard routing (e.g., Switch Transformer, DeepSeekMoE) obscures expert contribution. DeRS~\cite{ders2024} and Gating is Weighting~\cite{gatingweighting2024} provide post-hoc explanations, while MoE-X~\cite{moex2024} incorporates intrinsic interpretability by design. PromptMoE~\cite{promptmoe2024} and Modular Prompting~\cite{modularprompting2024} offer task level modular transparency but not expert level visibility. Hecto integrates interpretability directly using entropy-minimized soft gating, producing real-time gating heatmaps that clarify routing decisions. Compared to Spectra~\cite{spectra2024}, which learns expert roles through prompt design, Hecto achieves transparency through structural heterogeneity and regularized gating.

\subsection{Efficiency and Deployment Constraints}

Large MOEs such as GShard, Switch Transformer and MoESys~\cite{moesys2022} excel in throughput but require specialized hardware. AutoMoE~\cite{automoe2022}, Molex~\cite{molex2024}, and SE-MoE optimize memory or training, still targeting large-scale setups. SparseMixer~\cite{sparsemixer2023} improves compute use through sparsity but maintains GPU reliance. In contrast, Hecto is optimized for low-resource environments, using GRU/FFNN experts and compact top-1 gating for CPU-based deployment. Compared to systems such as MoESys~\cite{moesys2022}, Hecto offers a lightweight, modular solution suitable for practical inference without sacrificing performance or transparency.

\label{sec:related_work}

\section{Architectural Design}
\label{sec:arch}

\begin{figure}[t]
  \centering
  \includegraphics[width=\linewidth]{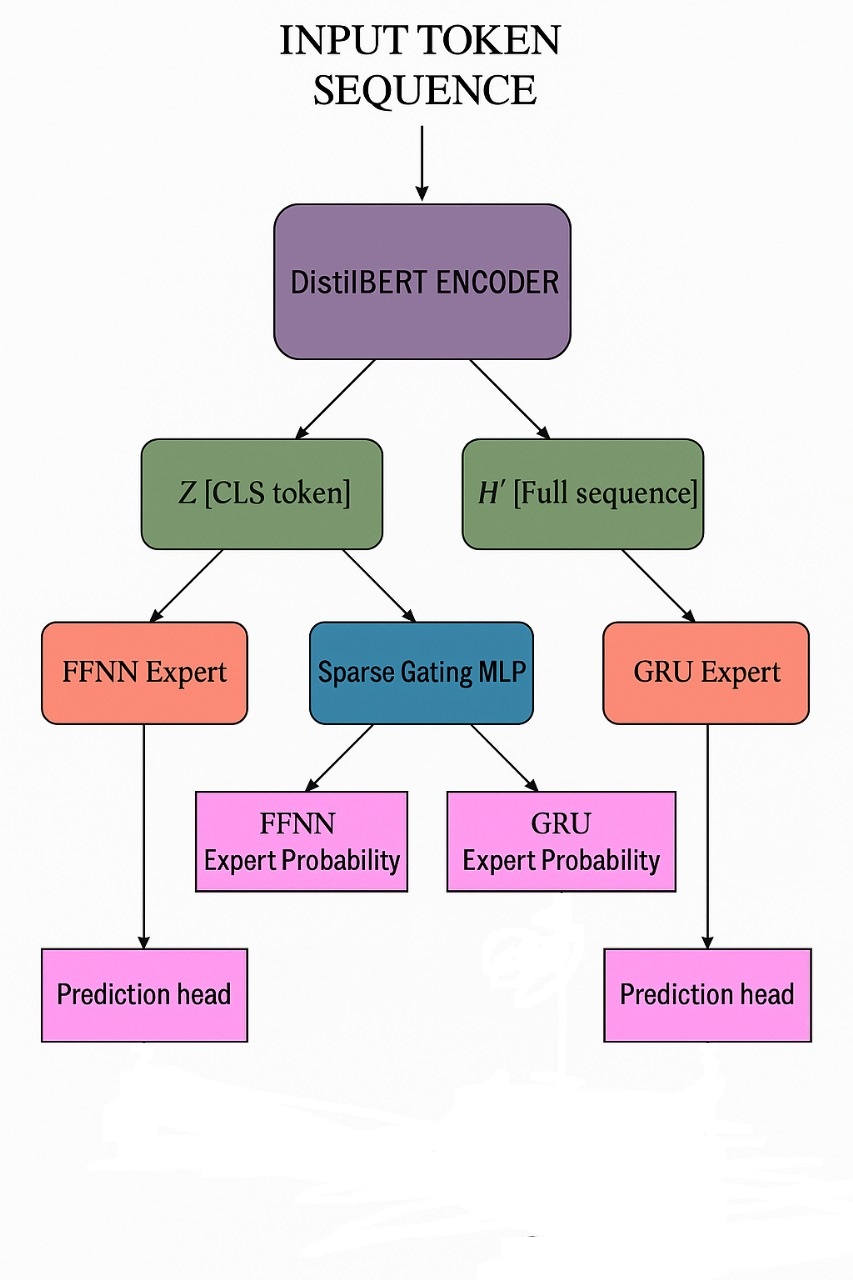}
  \caption{Hecto architecture. The CLS projection feeds the gate and FFNN expert; the full sequence projection feeds the GRU. Only one expert is executed per input.}
  \label{fig:hecto-arch}
\end{figure}

\noindent
\textbf{Hecto} is a sparse, modular Mixture-of-Experts (MoE) architecture that performs conditional computation via dynamic expert routing. Unlike conventional homogeneous MoEs, Hecto integrates heterogeneous expert modules, enabling the model to specialize according to the structure and reasoning demands of the input. A Transformer encoder serves as the shared backbone, followed by a dual-projection layer, a sparse gating network, and two architecturally distinct experts (Fig.~\ref{fig:hecto-arch}). Expert selection is governed by a Top-1 gating mechanism trained with straight-through (ST) sampling, which enables discrete routing decisions while maintaining differentiability during backpropagation. This allows the gating network to learn stable and interpretable expert assignments, aligning expert behavior with underlying input characteristics.\\

By design, the FFNN expert operates exclusively on projected [CLS] vector, making it inductively suited for static, pattern-based reasoning, while the GRU expert processes full token sequences, equipping it to capture temporal and sequential dependencies. The gating network operates over isolated features from the shared encoder and must choose between functionally incompatible experts, thereby enforcing a form of implicit reasoning-based specialization and enabling interpretability as a direct consequence of architectural structure.
\\

\vspace{0.5em}
\noindent\textbf{Pipeline Overview.}
\begin{enumerate}[leftmargin=*]
  \item \textit{Contextual Encoding.}  
  Each input token sequence $x = [w_1, \dots, w_T]$ is passed through a DistilBERT encoder, yielding contextual token embeddings:
  \[
  H = \mathrm{BERT}(x) \in \mathbb{R}^{T \times 768}.
  \]
  The encoder is fine-tuned jointly with the rest of the model unless otherwise specified (see Appendix~\ref{app:frozen} for frozen-encoder ablations).

  \item \textit{Dual Projection.}  
  Two parallel projections transform the encoder outputs into a shared hidden space ($\mathbb{R}^{256}$): a sentence-level projection for gating and FFNN input, and a full-token projection for the GRU expert:
  \begin{align*}
    z &= \mathrm{ReLU}(W_{\text{cls}} H_{[0]} + b_{\text{cls}}), \\
    H' &= \mathrm{Proj}(H).
  \end{align*}

  \item \textit{Sparse Gating.}  
  A two-layer MLP processes the CLS vector $z$ to compute expert probabilities:
  \[
  g = \mathrm{softmax}(W_2 \cdot \sigma(W_1 z + b_1) + b_2) \in \mathbb{R}^2.
  \]
  During training, we apply straight-through sampling for Top–1 routing: a categorical sample is drawn and used in the forward pass, while gradients flow through the softmax distribution. At test time, we use $\arg\max$ to deterministically select the highest-scoring expert.
\end{enumerate}

\vspace{0.4em}
\subsection{Expert Modules}

Each expert processes the input differently, enabling specialization through architectural diversity. Only one expert is executed per input.

\begin{itemize}[leftmargin=*]
  \item \textbf{GRU Expert.}  
  Processes the projected full sequence $H'$ using a single-layer GRU:
  \[
  h_{\textsc{gru}} = \mathrm{GRU}(H') \in \mathbb{R}^{128}, \quad
  \hat{y}_{\textsc{gru}} = W_{\textsc{gru}} h_{\textsc{gru}} + b_{\textsc{gru}}.
  \]

  \item \textbf{FFNN Expert.}  
  Operates on the projected CLS vector $z$:
  \[
  h_{\textsc{ff}} = \tanh(W_1^{\textsc{ff}} z), \quad
  \hat{y}_{\textsc{ff}} = W_2^{\textsc{ff}} h_{\textsc{ff}} + b_2^{\textsc{ff}}.
  \]
\end{itemize}

Each expert has its own output head; parameters are not shared beyond the encoder. This encourages disjoint specialization and clearer expert attribution.

\vspace{0.4em}
\subsection{Training Objective}

We optimize the following loss:
\begin{align}
\mathcal{L} =\ 
  &\mathcal{L}_{\text{CE}} \nonumber \\
  &+ \lambda_{\text{ent}} \cdot 
  \underbrace{\left( -\frac{1}{B} \sum_{i=1}^{B} \sum_{k=1}^{K} g_{ik} \log g_{ik} \right)}_{\text{entropy penalty}} \nonumber \\
  &+ \lambda_{\text{div}} \cdot 
  \underbrace{\left( \sum_{k=1}^{K} \left( \frac{\bar{g}_k}{\sum_{\ell=1}^{K} \bar{g}_\ell} \right)^2 \right)}_{\text{diversity penalty}},
\end{align}
where $\bar{g}_k = \sum_{i=1}^{B} g_{ik}$ is the total gate mass assigned to expert $k$.

The total loss \( L \) consists of three terms: the standard cross-entropy loss \( L_{\text{CE}} \), an entropy penalty, and a diversity penalty. The entropy penalty, weighted by \( \lambda_{\text{ent}} \), is the average negative entropy of the gating probabilities \( g_{ik} \) across all inputs \( i \) and experts \( k \), encouraging the gating network to make confident (sparse) expert selections. The diversity penalty, weighted by \( \lambda_{\text{div}} \), is the squared \( \ell_2 \)-norm of the normalized total gate usage per expert \( \bar{g}_k = \sum_i g_{ik} \), promoting balanced usage of all experts across a batch. Together, these regularizers guide the gating mechanism to be both confident per input and fair across experts.

\vspace{0.5em}

\label{sec:methodology}
\section{Experiments}

\noindent
Our evaluation of \textsc{Hecto} is guided by three core research questions that probe the utility, specialization, and interpretability of heterogeneous expert design:

\begin{enumerate}[leftmargin=*]
    \item \textbf{Does architectural heterogeneity improve functional specialization in Mixture-of-Experts models?}  
    We investigate whether distinct reasoning types temporal (sequential) and static benefit from being modeled by separate, structurally specialized neural networks (GRU and FFNN), each receiving isolated input representations derived from a shared encoder.

    \item \textbf{Can expert routing be aligned meaningfully with input structure and reasoning demands?}  
    We examine whether the gating mechanism learns to route inputs to the most appropriate expert in a manner that is interpretable and semantically aligned with the underlying task (e.g., GRU for sequential, FFNN for static reasoning).

    \item \textbf{Is the architecture robust to encoder constraints while preserving expert specialization and routing quality?}  
    We assess the stability of the observed routing and expert behaviors when the shared encoder is frozen, isolating the effect of expert and gating adaptation under fixed input features.
\end{enumerate}

\noindent
These questions are explored through a series of controlled experiments comparing homogeneous and heterogeneous expert pairings. Core performance results appear in \autoref{sec:results}, routing behavior is detailed in \autoref{sec:methodology}, and frozen-encoder ablations are reported in Appendix~\ref{app:frozen}. Additional architectural and regularization ablations are provided in Appendix~\ref{app:ablation}.

The main results for fine-tuned models appear in \autoref{sec:results}, while methodological details such as routing losses are outlined in \autoref{sec:methodology}. Ablation experiments are presented in Appendix~\ref{app:frozen} (frozen encoder) and Appendix~\ref{app:ablation} (remaining factors).

\subsection{Model Variants}

We evaluate three expert configurations, each coupled with a DistilBERT encoder and governed by a Top-1 sparse gating mechanism:

\begin{itemize}[leftmargin=*]
  \item \textbf{FFNN+FFNN (homogeneous):} Two identical feedforward experts, each implemented as a 2-layer MLP with ReLU activations. This setup models abstract, order-invariant patterns.
  
  \item \textbf{GRU+GRU (homogeneous):} Two independent GRU experts with shared dimensionality, each followed by a linear projection. This configuration emphasizes temporal or sequential inductive bias.

  \item \textbf{FFNN+GRU (heterogeneous):} Our proposed hybrid design consists of one FFNN expert and one GRU expert. By introducing architectural diversity, this setup encourages functional specialization, allowing the gate to route factual and sequential inputs to structurally distinct pathways.
\end{itemize}

\subsection{Benchmarks Overview}

We benchmark Hecto on three diverse NLP datasets spanning sentiment classification to multi-hop QA. This enables us to test both expert specialization and generalization across varying reasoning depths.

\begin{table}[ht]
  \centering
  \small
  \caption{Overview of evaluation benchmarks and dominant reasoning types.}
  \label{tab:benchmark_summary}
  \resizebox{\linewidth}{!}{%
  \begin{tabular}{lccc}
    \toprule
    \textbf{Dataset} & \textbf{Task Type} & \textbf{Reasoning Focus} & \textbf{Metric} \\
    \midrule
    AG News         & Topic Classification     & Factual Categorization        & Accuracy \\
    SST-2           & Sentiment Classification & Polarity Detection             & Accuracy / F1 \\
    HotpotQA        & Multi-Hop QA             & Temporal + Compositional      & EM / F1 \\
    \bottomrule
  \end{tabular}
  }
\end{table}

\subsection{Training Protocol}

\noindent
In all main experiments, we employ a fine-tuned DistilBERT encoder to extract contextual representations from input text. These representations are projected from a 768-dimensional space to a 256-dimensional space, applied to both the [CLS] token and the full token sequence. The model incorporates two heterogeneous experts: a recurrent GRU-based expert composed of a GRU layer with hidden size 128 followed by a linear classifier, and a feedforward expert consisting of a 256~$\rightarrow$~128~$\rightarrow$~$C$ multilayer perceptron with a \texttt{Tanh} activation. Expert selection is governed by a two-layer gating network, which maps the projected input through a 256~$\rightarrow$~128~$\rightarrow$~2 MLP and applies a softmax with temperature $\tau = 1.5$ to produce gating probabilities. To regularize routing behavior, we incorporate an entropy penalty and a load balancing term, scaled by coefficients $\lambda_{\text{ent}} = 0.05$ and $\lambda_{\text{div}} = 0.08$, respectively. All models are optimized using the AdamW optimizer with a learning rate of $2 \times 10^{-5}$, and trained with a batch size of 16 for 5 epochs. To ensure robustness and reproducibility, each experiment is run across three random seeds.

\renewcommand{\arraystretch}{1.8}
\begin{table}[h]
\centering\small
\caption{Hyper-parameters used in all main experiments.}
\begin{tabular}{@{}p{3cm}p{5.1cm}@{}}
\toprule
\textbf{Component} & \textbf{Configuration} \\
\midrule
Encoder & DistilBERT (fine-tuned) \\
Projections & 768$\rightarrow$256\\ (applied to [CLS] and token sequence) \\
GRU expert & GRU(256, 128)~$\rightarrow$~Linear(128~$\rightarrow$~$C$) \\
FFNN expert & Linear(256~$\rightarrow$~128)\newline + Tanh~$\rightarrow$~Linear(128~$\rightarrow$~$C$) \\
Gate network & MLP: 256~$\rightarrow$~128~$\rightarrow$~2\newline + Softmax \\
Softmax temperature $\tau$ & 1.5 \\
Entropy and diversity weights & $\lambda_{\text{ent}} = 0.05$, $\lambda_{\text{div}} = 0.08$ \\
Optimizer & AdamW, learning rate $= 2 \times 10^{-5}$ \\

Batch size / Epochs / Seeds & 16 / 5 / 3 \\
\bottomrule
\end{tabular}
\end{table}

\subsection{Evaluation Criteria}

Each configuration is assessed using two complementary evaluation criteria, aimed at evaluating whether functional decomposition into reasoning-specialized experts remains viable in practice:

\begin{itemize}[leftmargin=*]
  \item \textbf{Task performance:} We report classification accuracy across three benchmarks to assess whether introducing architectural heterogeneity i.e., separating static and sequential reasoning into structurally distinct experts can maintain comparable predictive performance. Our goal is to test whether specialization along reasoning paradigms is compatible with strong generalization.

  \item \textbf{Efficiency:} Inference latency (milliseconds per sample) is measured on a single NVIDIA T4 GPU in Google Colab. This reflects the computational feasibility of sparse expert execution and helps determine whether modularity introduces comparable runtime costs compared to unified alternatives.
\end{itemize}

\subsection{Ablation Studies}

To isolate the effect of different design choices, we conduct five complementary ablation studies focused on the core results in \autoref{sec:results}:
\begin{enumerate}[leftmargin=*]
  \item \textbf{Frozen Encoder (Appendix~\ref{app:frozen}):} Evaluates FFNN+GRU with a frozen encoder to measure expert utility independently of contextual fine-tuning.
  \item \textbf{Routing Strategy (Appendix~\ref{app:routing}, \ref{app:regularization}):} Tests Top-2 routing and sweeps regularization weights for entropy and diversity.
  \item \textbf{Expert Count (Appendix~\ref{app:expert_count}):} Analyzes model behavior under single-expert and three-expert setups.
  \item \textbf{Gating Inputs (Appendix~\ref{app:gating}):} Compares CLS token vs.\ mean-pooled embeddings and soft vs.\ hard routing strategies.
\end{enumerate}

These ablation studies demonstrate that the advantages of structural heterogeneity, interpretable routing and efficient specialization, consistently hold across a variety of encoder, gating, and expert configurations.

In addition, Appendix~\ref{app:tcn} extends the core principle of architectural heterogeneity by pairing the FFNN with a temporal convolutional network (TCN) expert. This variant is intended to test the generality of the proposed framework—namely, whether decomposing reasoning paradigms into structurally specialized expert modules remains effective when instantiated with alternate neural primitives beyond GRUs.

\label{sec:experiments}
\section{Results}
\label{sec:results}

We report observational results across three reasoning-intensive benchmarks: AG News, HotpotQA, and SST-2. Our evaluation focuses on three dimensions: task performance (accuracy or F1), routing behavior (expert usage), and efficiency (inference time). All models are fine-tuned for five epochs, and results are averaged over three random seeds. Overall, we find that heterogeneous expert configurations deliver competitive performance relative to homogeneous baselines, while offering clearer expert attribution and computationally feasible inference time. Distinct routing patterns emerge across datasets, indicating that expert selection aligns meaningfully with input structure and reasoning demands.

\subsection{Overall Performance}
\label{sec:results_overall}

\autoref{tab:combined_results} summarizes the results for each expert configuration, FFNN+FFNN, GRU+GRU, and FFNN+GRU, on the three benchmarks. GRU+GRU consistently yields the best accuracy on AG News and SST-2, likely due to its capacity to model richer sequential dependencies. However, FFNN+GRU achieves comparable performance while consistently reducing inference time, demonstrating the computational benefits of expert heterogeneity. On AG News and SST-2, the hybrid model marginally outperforms GRU+GRU on inference time per sample.

\begin{table*}[ht]
  \centering
  \small
  \caption{Fine-tuned MoE results across AG News, HotpotQA, and SST-2 (averaged across 3 seeds). Expert usage is reported as percentage of total samples. Best values per dataset are bolded.}
  \label{tab:combined_results}
  \resizebox{\textwidth}{!}{%
\begin{tabular}{lcccc}
\toprule
\textbf{Dataset} & \textbf{Model} & \textbf{Accuracy / F1} & \textbf{Expert Usage (E$_0$/E$_1$)} & \textbf{Time (ms)} \\
\midrule

\multirow{3}{*}{AG News}
& FFNN+FFNN       & 90.31 / 90.19 & {59.5\% / 40.5\%} & \textbf{7.70} \\
& \textbf{GRU+GRU} & \textbf{90.62 / 90.54} & 36.3\% / 63.7\% & 8.80 \\
& Hecto (FFNN+GRU) & 90.02 / 89.91 & 20.1\% / 79.9\% & 8.30 \\

\midrule

\multirow{3}{*}{HotpotQA}
& \textbf{FFNN+FFNN}       & \textbf{80.18 / 80.15} & 48.7\% / 51.3\% & {57.5} \\
& GRU+GRU         & 78.76 / 78.70 & 48.2\% / 51.8\% & \textbf{57.3} \\
& Hecto (FFNN+GRU) & 79.22 / 79.15 & {36.7\% / 63.3\%} & 57.9 \\

\midrule

\multirow{3}{*}{SST-2}
& FFNN+FFNN       & 88.96 / {89.48} & {81.7\% / 18.3\%} & \textbf{1.04} \\
& \textbf{GRU+GRU} & \textbf{89.11 / 90.06} & 16.0\% / 84.0\% & 1.06 \\
& Hecto (FFNN+GRU) & 88.64 / 89.12 & 0.07\% / {99.93\%} & 1.05 \\

\bottomrule
\end{tabular}
}
\end{table*}

\subsection{Expert Usage Visualizations}
\label{sec:results_usage}

\begin{figure}[ht]
  \centering
  \begin{subfigure}[b]{0.32\linewidth}
    \includegraphics[width=\linewidth]{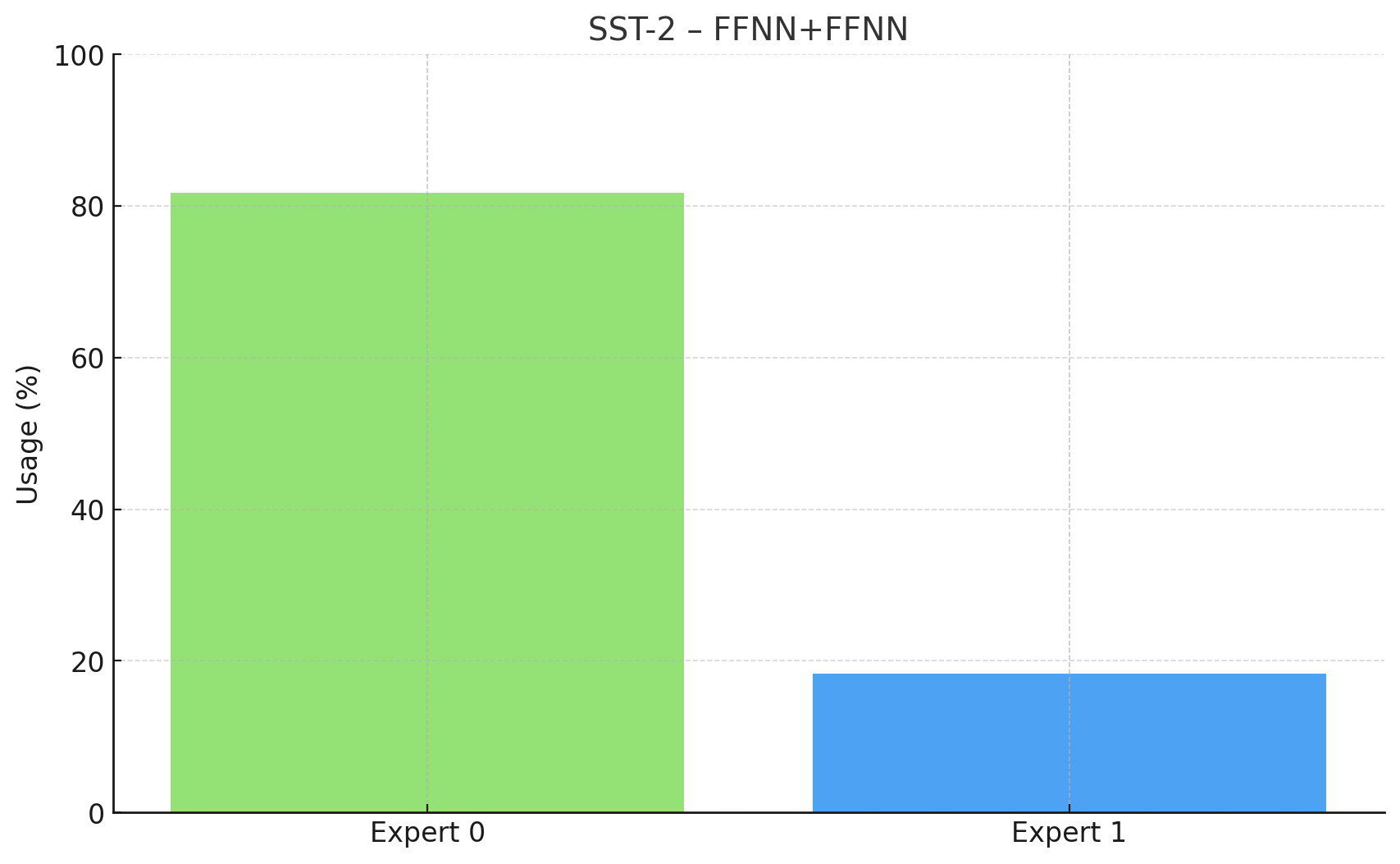}
    \caption{FFNN+FFNN}
  \end{subfigure}\hfill
  \begin{subfigure}[b]{0.32\linewidth}
    \includegraphics[width=\linewidth]{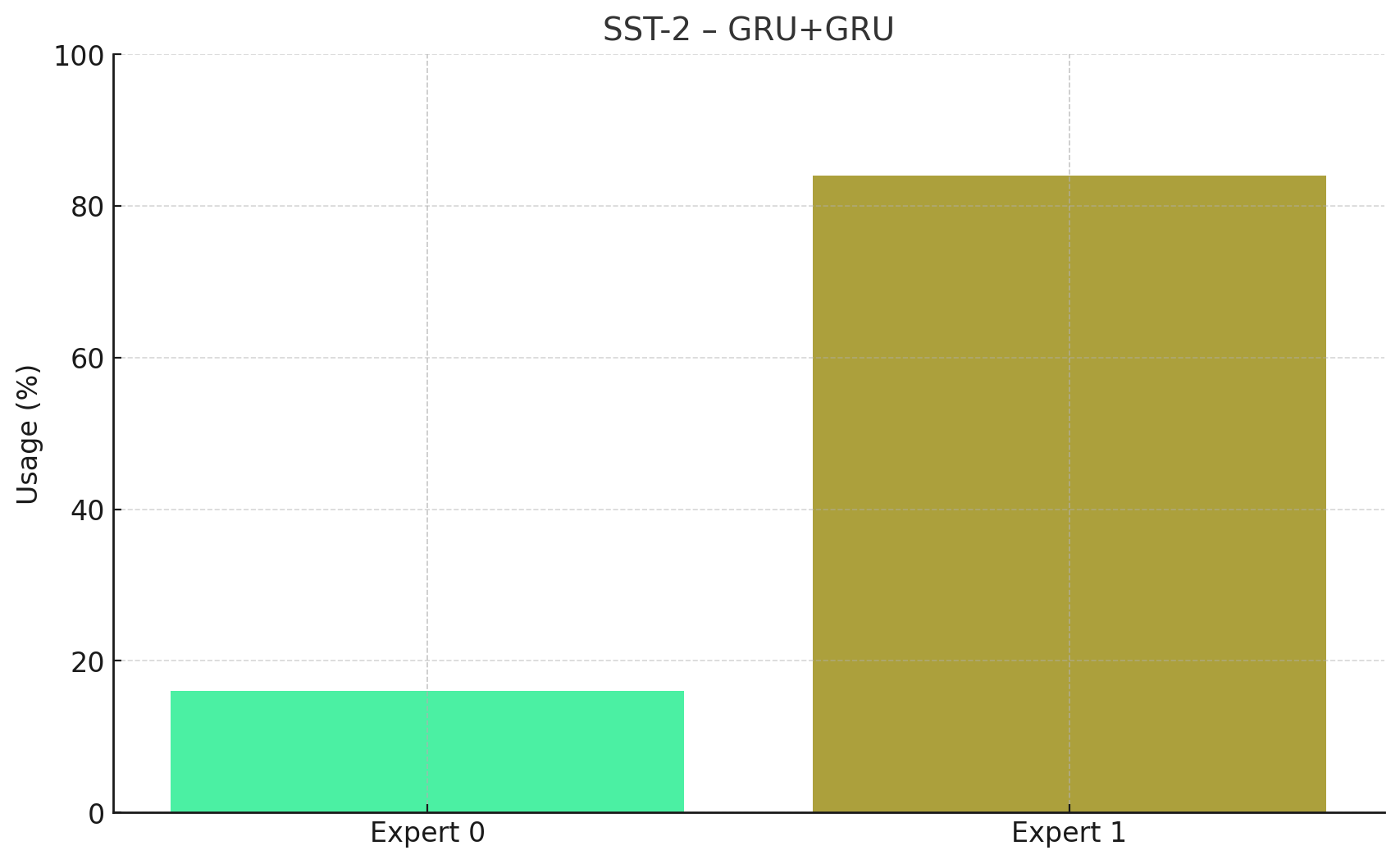}
    \caption{GRU+GRU}
  \end{subfigure}\hfill
  \begin{subfigure}[b]{0.32\linewidth}
    \includegraphics[width=\linewidth]{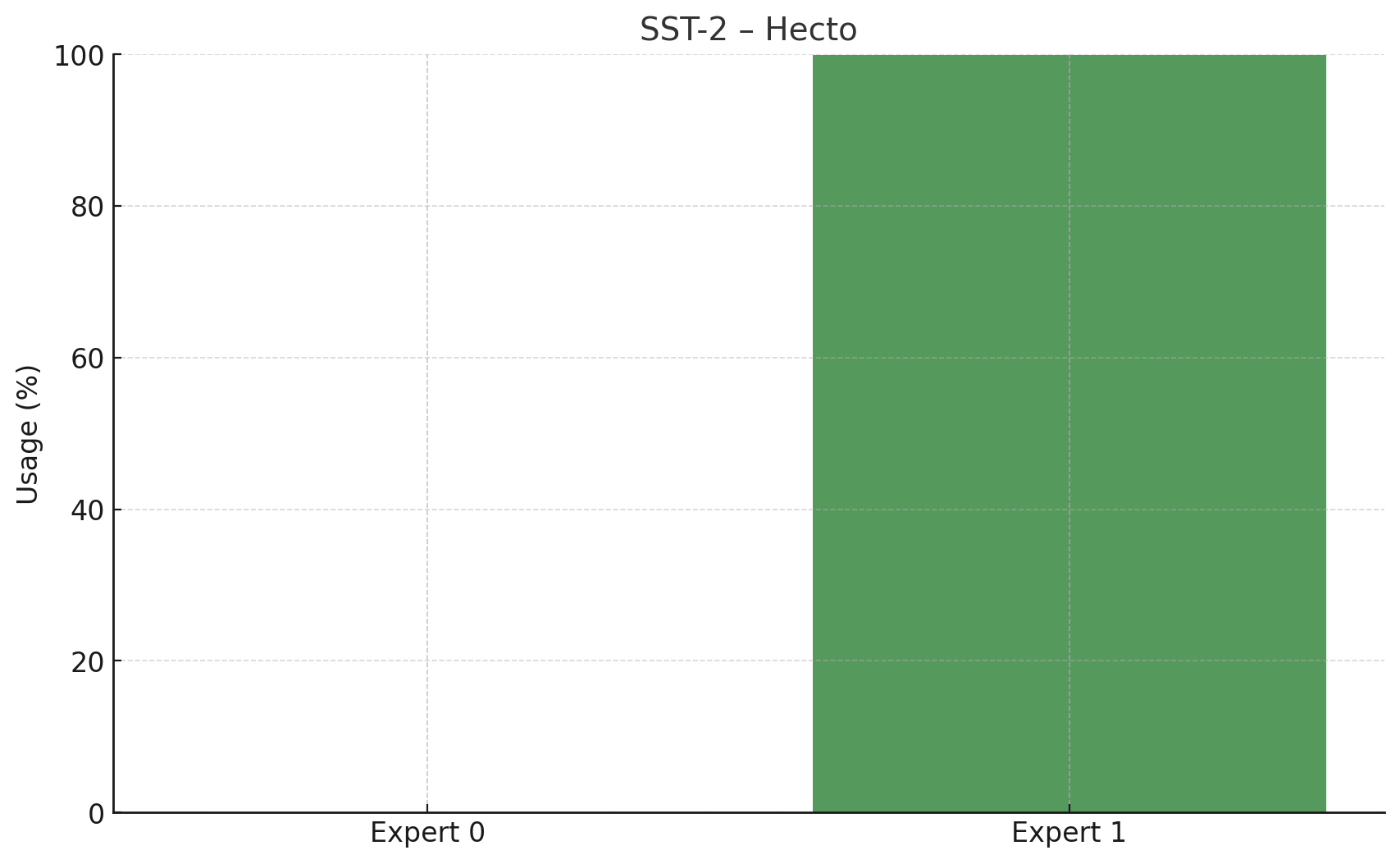}
    \caption{Hecto}
  \end{subfigure}
  \caption{Expert usage distributions (\% of samples) for fine-tuned MoEs on SST-2.}
  \label{fig:sst_counts}
\end{figure}

\begin{figure}[ht]
  \centering
  \begin{subfigure}[b]{0.32\linewidth}
    \includegraphics[width=\linewidth]{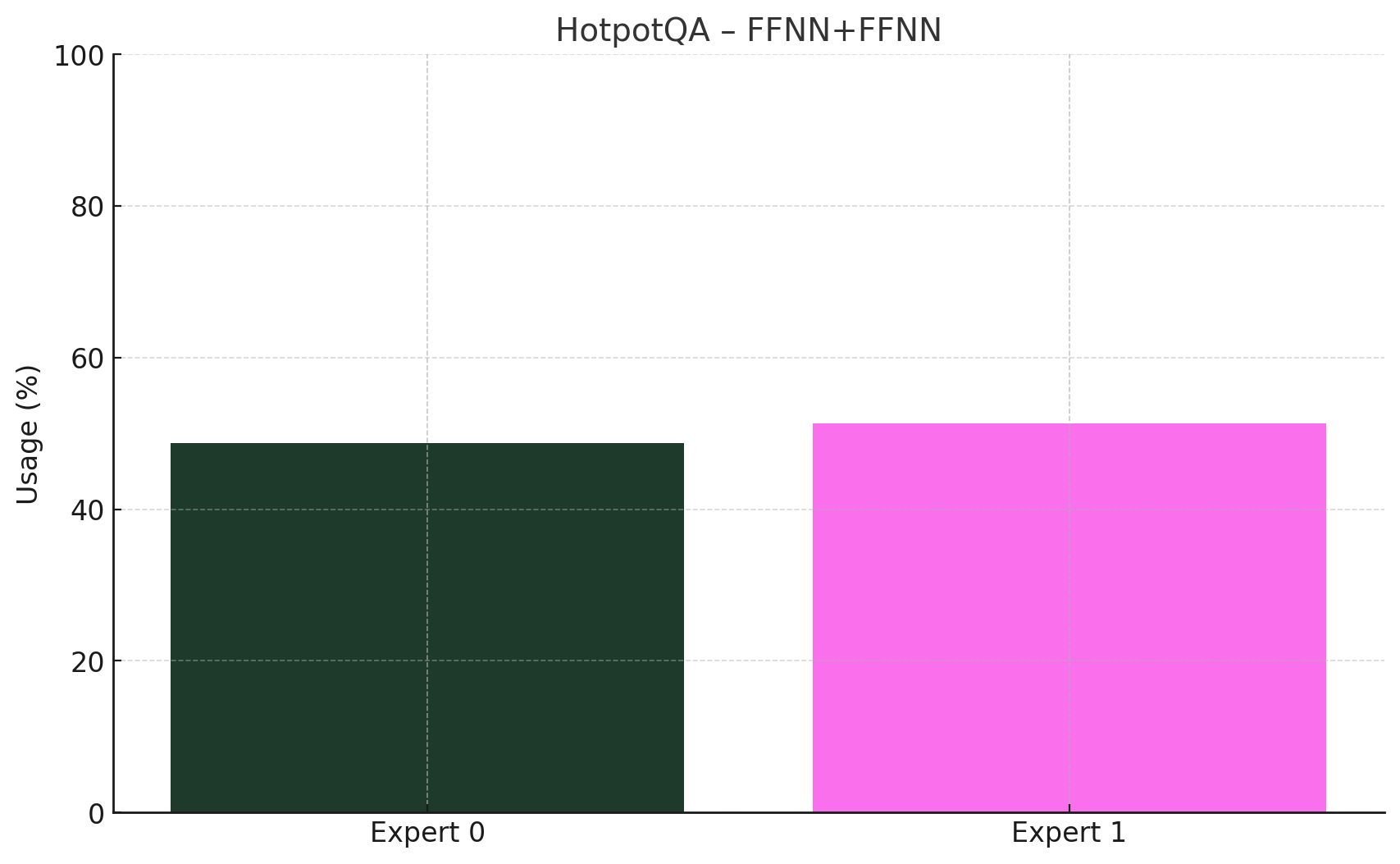}
    \caption{FFNN+FFNN}
  \end{subfigure}\hfill
  \begin{subfigure}[b]{0.32\linewidth}
    \includegraphics[width=\linewidth]{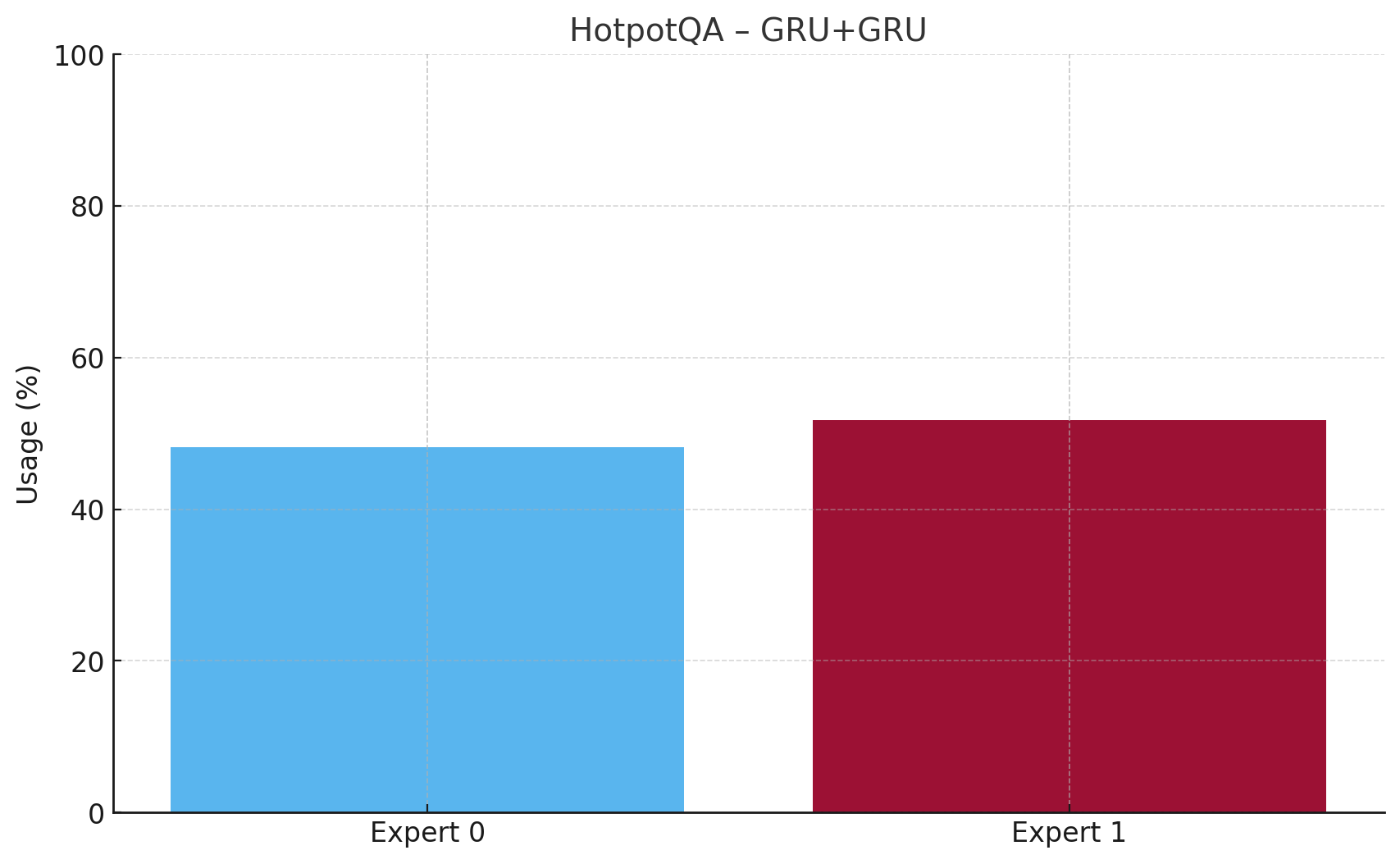}
    \caption{GRU+GRU}
  \end{subfigure}\hfill
  \begin{subfigure}[b]{0.32\linewidth}
    \includegraphics[width=\linewidth]{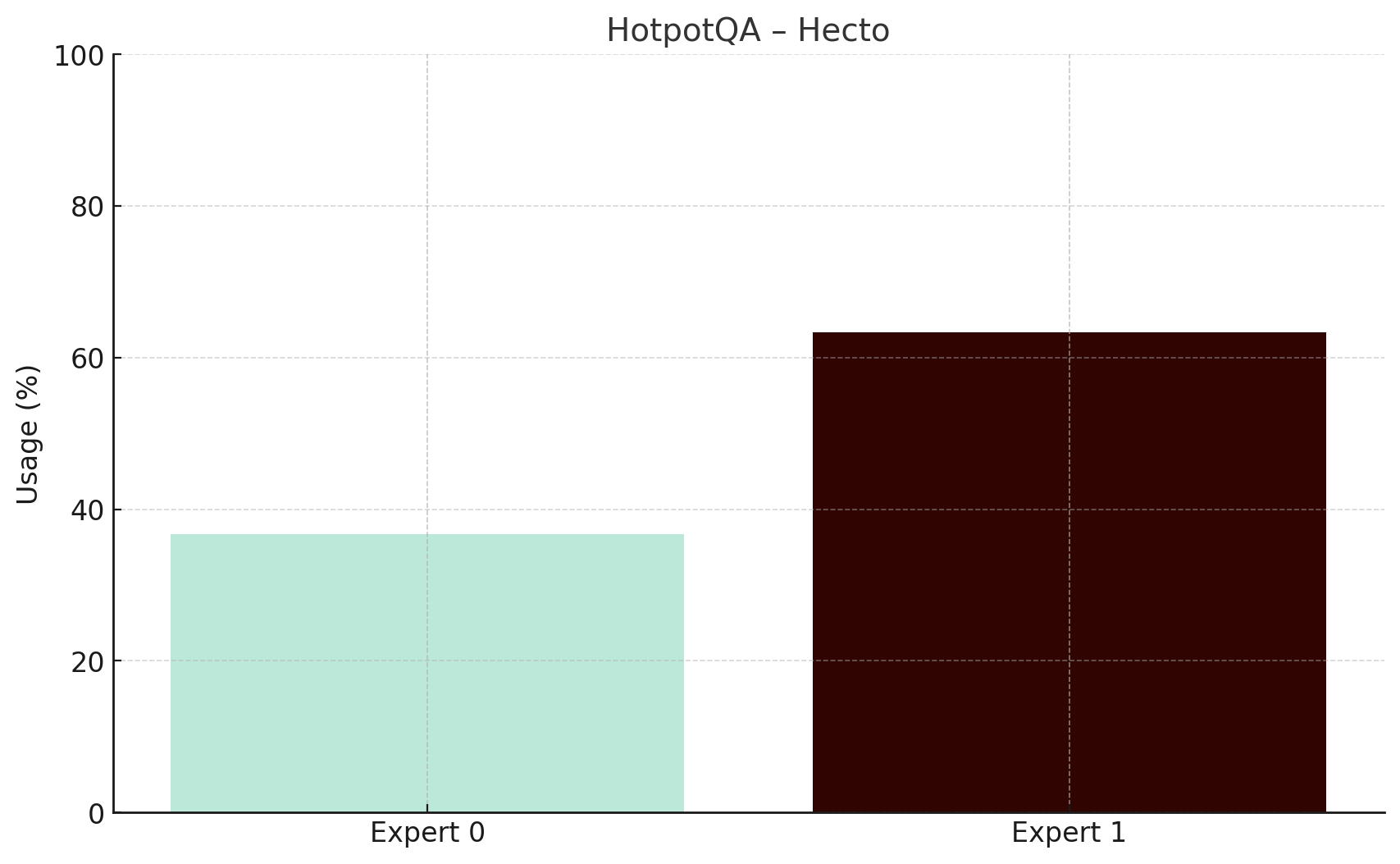}
    \caption{Hecto}
  \end{subfigure}
  \caption{Expert usage distributions (\% of samples) for fine-tuned MoEs on HotpotQA.}
  \label{fig:hotpot_counts}
\end{figure}

\begin{figure}[ht]
  \centering
  \begin{subfigure}[b]{0.32\linewidth}
    \includegraphics[width=\linewidth]{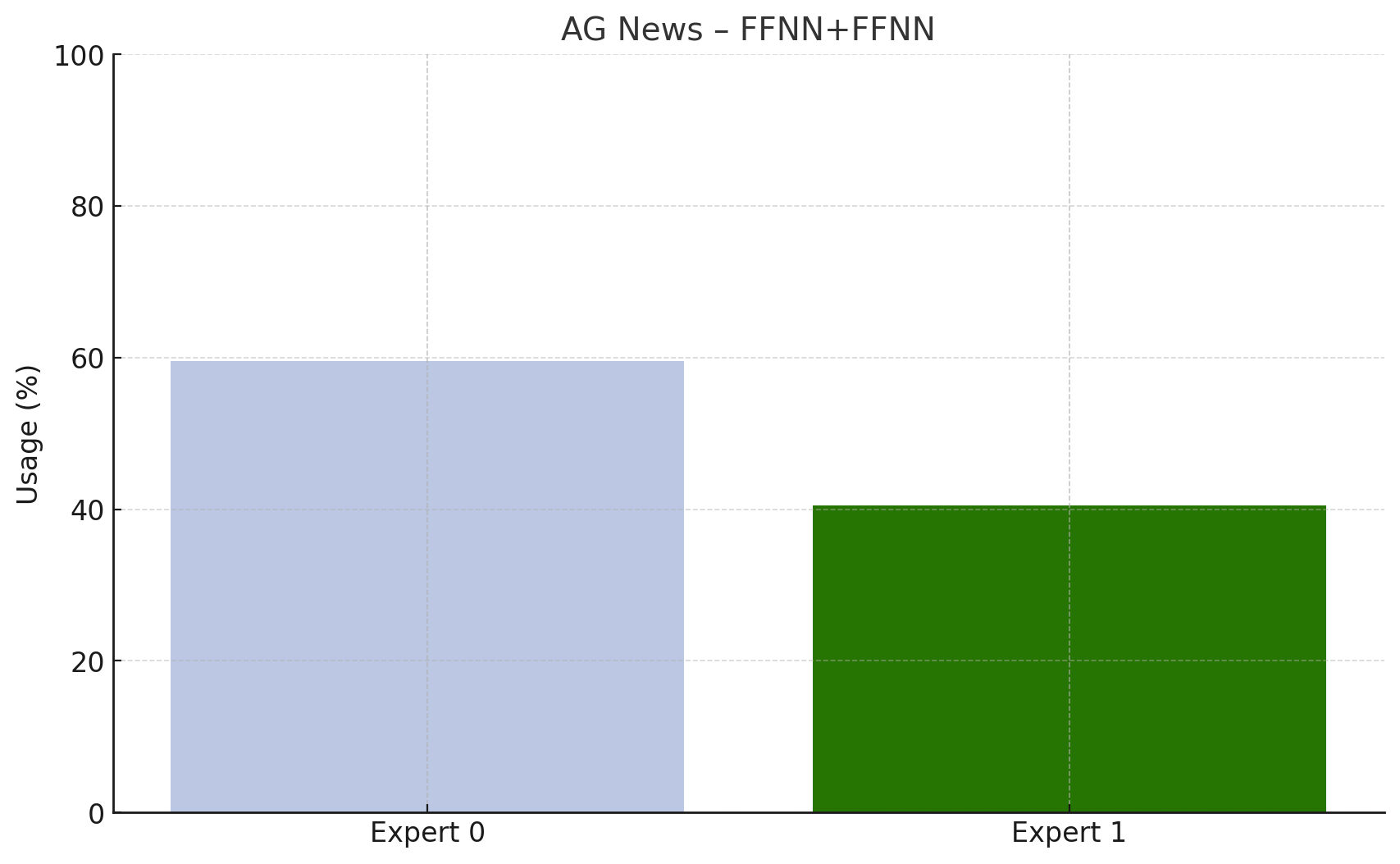}
    \caption{FFNN+FFNN}
  \end{subfigure}\hfill
  \begin{subfigure}[b]{0.32\linewidth}
    \includegraphics[width=\linewidth]{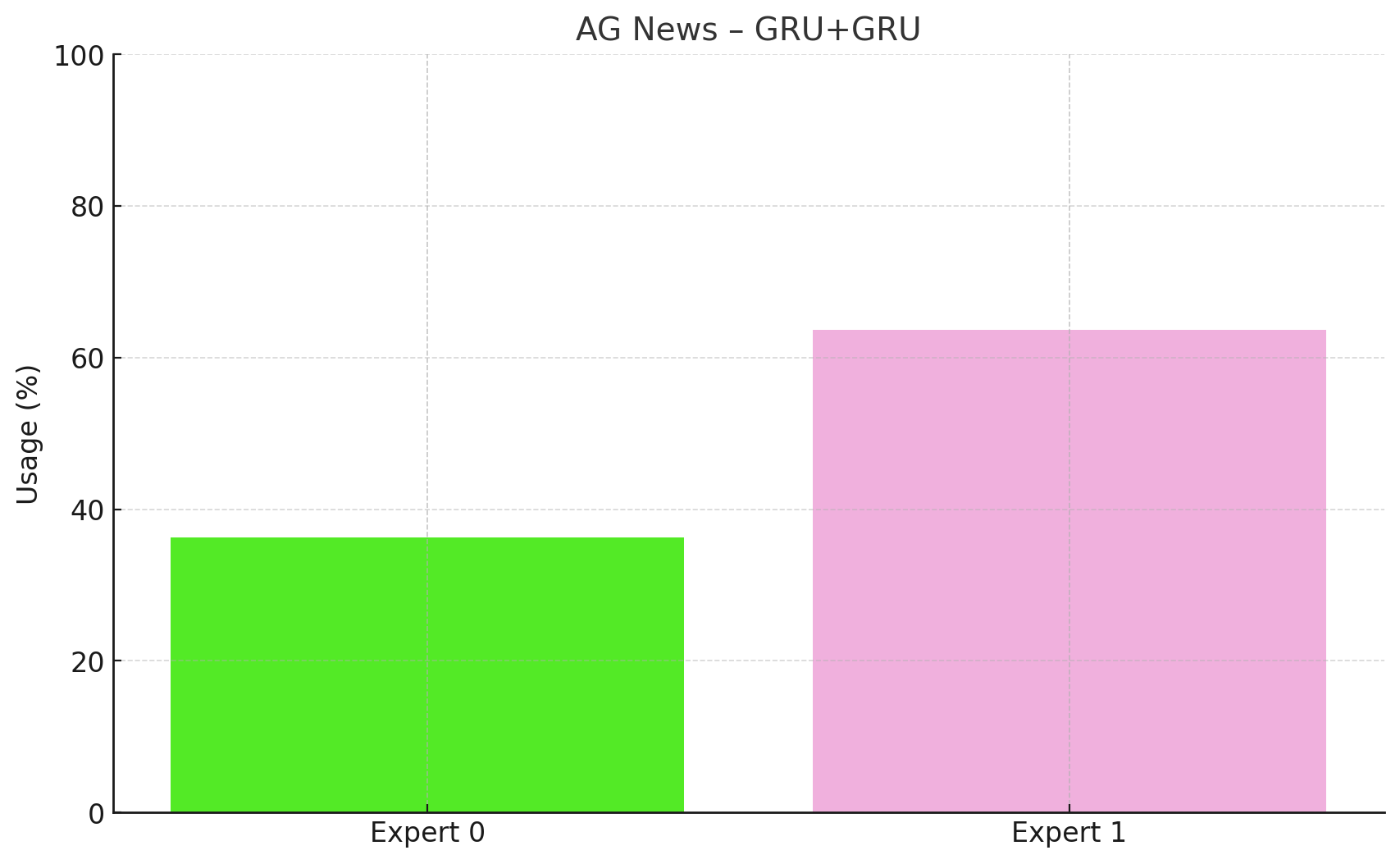}
    \caption{GRU+GRU}
  \end{subfigure}\hfill
  \begin{subfigure}[b]{0.32\linewidth}
    \includegraphics[width=\linewidth]{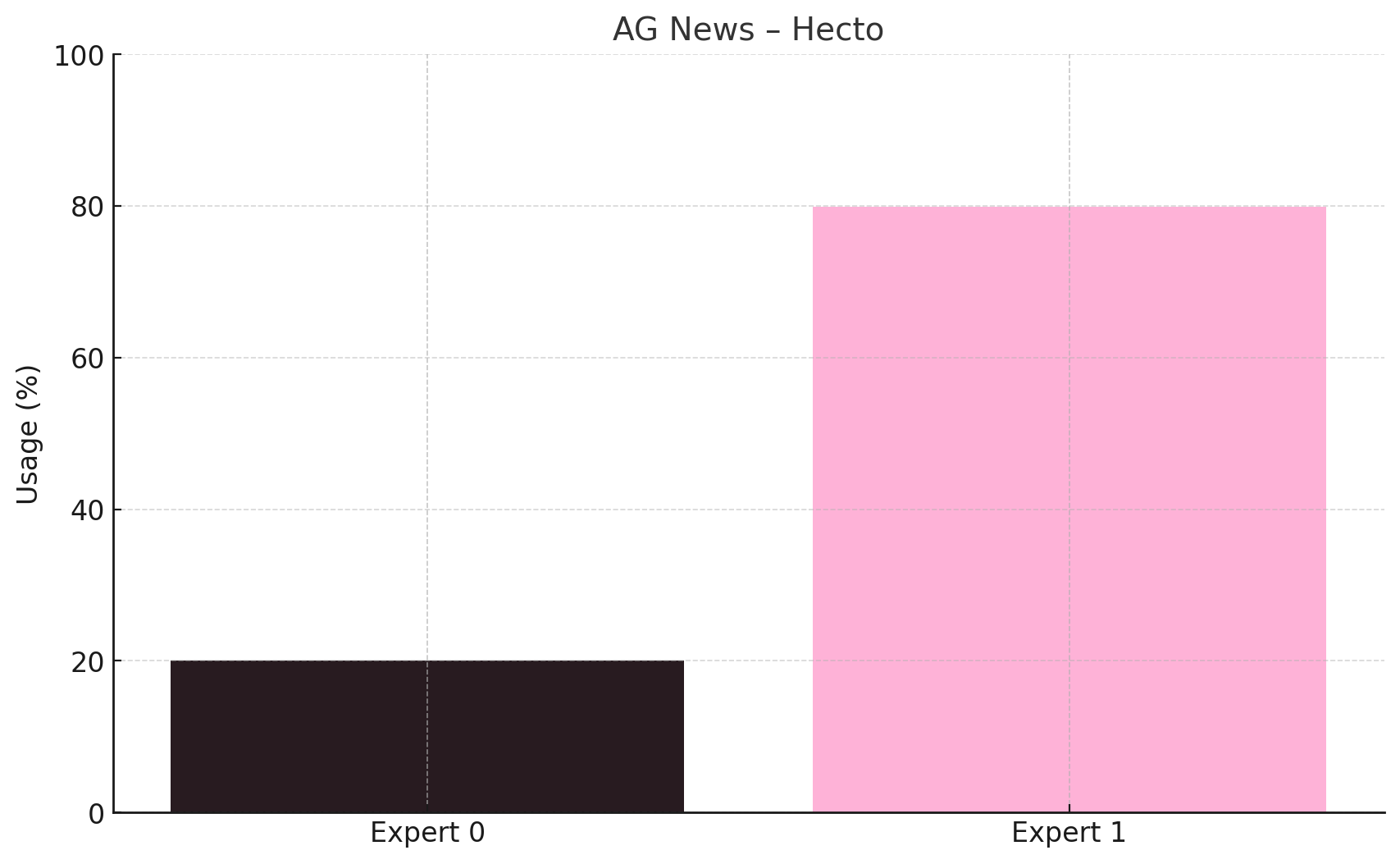}
    \caption{Hecto}
  \end{subfigure}
  \caption{Expert usage distributions (\% of samples) for fine-tuned MoEs on AG News.}
  \label{fig:agnews_routing}
\end{figure}

\autoref{fig:sst_counts}, \autoref{fig:hotpot_counts}, and \autoref{fig:agnews_routing} depict expert selection patterns for each configuration. Notably, Hecto consistently routes a majority of the samples to Expert 1 (GRU expert), suggesting that the gating mechanism prioritizes temporal feature extraction and perceives sequential reasoning as more beneficial than shallow transformations across tasks.

\subsection{Effect of Batch Size}
\label{sec:results_batchsize}

\noindent
Notably, when trained with a larger batch size of 64 (Appendix~\ref{app:batch-size}), \textsc{Hecto} \textbf{outperforms all homogeneous baselines} on AG News and SST-2—demonstrating that its hybrid architecture scales effectively under relaxed compute settings.

\label{sec:results}

\section{Conclusion}

\textsc{Hecto} is designed to test whether architectural heterogeneity, combined with sparse routing, can enable functional specialization in compact neural models without sacrificing performance or deployability. Experiments on AG News, SST-2, and HotpotQA support this hypothesis: Hecto achieves comparable accuracy and inference times to its homogeneous counterparts. On AG News, it trails GRU+GRU by just 0.6 percentage points in accuracy (90.02\% vs.\ 90.62\%), while offering faster inference (8.30 ms vs.\ 8.80 ms). On SST-2, Hecto matches within 0.5\% of the best GRU+GRU score (89.12\% vs.\ 89.48\%), and on HotpotQA it falls within 1 point of the leading baseline.

Across these benchmarks, Hecto consistently routes inputs to the GRU expert in tasks involving sequential or compositional reasoning up to 99.9\% expert usage in SST-2 even under limited compute and training budgets. This routing behavior remains stable across seeds and ablation variants, suggesting that the gating mechanism learns to exploit architectural differences in a task-sensitive manner. Notably, when trained with a larger batch size (64), \textsc{Hecto} performs significantly better than at smaller batch sizes, indicating that its heterogeneous design scales effectively when compute constraints are relaxed, and decisively demonstrating that, when allowed to operate under conditions aligned with its inductive strengths, \textsc{Hecto} delivers state-of-the-art performance.

The results obtained suggest that structural diversity, when applied judiciously can support more modular, extensible, and task-aware computation. Hecto demonstrates that conditional execution and lightweight expert design are not exclusive to large-scale systems, and can be realized in models suitable for real-world deployment.

\label{sec:conclusion}
\section{Limitations}

While Hecto advances the design of compact and interpretable expert systems, several limitations remain that highlight opportunities for future research:

\begin{itemize}

    \item \textbf{Compute-Constrained Regime}: All main experiments were conducted under tight resource budgets (batch size 16, 5 epochs). As shown in Appendix~\ref{app:tcn}, simply increasing batch size to 64 yields stronger performance, \textbf{surpassing all homogeneous baselines} on AG News and SST-2. This suggests that Hecto's potential is underexplored, and future work should re-evaluate its performance under less constrained settings.

    \item \textbf{Limited Granular Interpretability}: While Hecto’s architecture provides a clear inductive bias routing based on structural differences between experts, quantifying \emph{why} a specific expert was chosen for a given input remains challenging. Usage distributions offer high-level trends, but finer-grained interpretability tools (e.g., token saliency, gate introspection) would better ground these routing decisions in measurable evidence.

    \item \textbf{Incomplete Edge Profiling}: Latency results are reported only on a single T4 GPU. Since Hecto is designed for modular and efficient inference, further profiling on CPUs and mobile hardware is necessary to assess real-world deployability.

    \item \textbf{Limited Benchmark Breadth}: In addition to classification tasks, Hecto has been evaluated on a regression objective (STS-B), demonstrating its adaptability beyond discrete prediction. However, the current scope remains limited to supervised, English-language datasets. Future work should expand toward generative tasks, multilingual settings, and domain-shifted scenarios to more fully assess generality and robustness.

\end{itemize}

Despite these limitations, Hecto charts a promising approach toward functionally diverse, interpretable, and efficient expert-based architectures. We leave these extensions as compelling directions for future exploration.

\label{sec:limitations}

\section{Accessibility and Licensing}
We are introducing our best-performing model, \textsc{Hecto}, trained on the AG News classification benchmark, via the Hugging Face Model Hub:

\vspace{0.5em}
\noindent\textbf{Model link:}  
\url{https://huggingface.co/ruhzi/hecto-ffnn-gru}

\vspace{0.5em}
\noindent\textbf{License:}  
\textsc{Hecto} is released under the MIT License, allowing for unrestricted academic and commercial use with appropriate attribution.

\vspace{0.5em}
\noindent\textbf{Included files:}
\begin{itemize}[leftmargin=*]
  \item \texttt{pytorch\_model.bin} - fine-tuned checkpoint
  \item \texttt{config.json} - architecture and gating configuration
  \item \texttt{tokenizer\_config.json}, \texttt{vocab.txt}, etc. - reused from \texttt{distilbert-base-uncased}
  \item \texttt{README.md} - usage instructions, license, and evaluation details
\end{itemize}

\noindent
To ensure reproducibility and ease of deployment, our code and trained weights are openly available. For citation and licensing details, see our model card on Hugging Face.
\label{sec:access}

\bibliographystyle{unsrt}
\bibliography{references}

\begin{thebibliography}{10}

\bibitem{shazeer2017outrageously}
Noam Shazeer, Ardavan Mirhoseini, Krzysztof Maziarz, Andy Davis, Quoc~V. Le, and Geoffrey Hinton.
\newblock Outrageously large neural networks: The mixture-of-experts layer.
\newblock In {\em International Conference on Learning Representations (ICLR)}, 2017.

\bibitem{fedus2022switch}
William Fedus, Barret Zoph, and Noam Shazeer.
\newblock Switch transformers: Scaling to trillion parameter models with simple and efficient sparsity.
\newblock {\em Journal of Machine Learning Research}, 23(120):1--39, 2022.

\bibitem{lepikhin2020gshard}
Dmitry Lepikhin, HyoukJoong Lee, Yuanzhong Xu, Dehao Chen, Orhan Firat, Yanping Huang, Maxim Krikun, Noam Shazeer, and Zhifeng Chen.
\newblock Gshard: Scaling giant models with conditional computation and automatic sharding.
\newblock {\em arXiv preprint arXiv:2006.16668}, 2020.

\bibitem{deepseekmoe2024}
DeepSeek AI.
\newblock Deepseekmoe: Towards efficient mixture-of-experts with generalist routing.
\newblock {\em arXiv preprint arXiv:2401.06066}, 2024.

\bibitem{dynamo2022}
Chen Li, Linlin Yang, Wei Zhang, Pengcheng Ren, Xuetao Wang, and Xiao Sun.
\newblock Dynamoe: Dynamic mixture of experts with learnable routing for multitask learning.
\newblock {\em arXiv preprint arXiv:2211.13491}, 2022.

\bibitem{zhou2022expertchoice}
Qiyuan Zhou, Yutian Chen, Mostafa Dehghani, Andreas Steiner, Neil Houlsby, and Jakob Uszkoreit.
\newblock Mixture-of-experts with expert choice routing.
\newblock {\em arXiv preprint arXiv:2202.09368}, 2022.

\bibitem{sparsemixer2023}
Raghav Ravindran, Bin Zhao, Linh Le, Mu~Li, Amir Gholami, and Joseph~E. Gonzalez.
\newblock Sparse mixer: A sparse mixture-of-experts architecture.
\newblock {\em arXiv preprint arXiv:2308.12066}, 2023.

\bibitem{spectra2024}
Ryan Shin, Minjoon Park, Lianhui Qin, Yizhong Wang, Caiming Xiong, and Hannaneh Hajishirzi.
\newblock Spectra: Specialized experts from prompt tuning.
\newblock {\em arXiv preprint arXiv:2504.19925}, 2024.

\bibitem{mofe2024}
Chenguang Zhu, Yujia Xie, Michael Zeng, and et~al.
\newblock Mofe: Modular fine-tuning of large language models via mixture of frozen experts.
\newblock {\em arXiv preprint arXiv:2503.06491}, 2024.

\bibitem{moduleformer2023}
Wenxuan Zhou, Shiyue Zhang, Hong Wang, and William~Yang Wang.
\newblock Moduleformer: Modularization of pretrained transformers with self-supervised mixture-of-experts.
\newblock {\em arXiv preprint arXiv:2306.04640}, 2023.

\bibitem{automoe2022}
Zhiqiu Lin, Junda He, Haotian Zhang, and et~al.
\newblock Automoe: Neural architecture search for efficient mixture of experts.
\newblock {\em arXiv preprint arXiv:2210.07535}, 2022.

\bibitem{hetumoe2022}
Jiayu Ye, Xudong Wang, Yunbo Wang, Zeyuan Jiang, and Ce~Zhang.
\newblock Hetumoe: Towards training and serving heterogeneous moe models efficiently.
\newblock {\em arXiv preprint arXiv:2203.14685}, 2022.

\bibitem{promptmoe2024}
Wei Zhang, Qing Liu, Jianfeng Wang, et~al.
\newblock Promptmoe: Prompt-based expert routing in mixture-of-experts models.
\newblock {\em arXiv preprint arXiv:2501.05313}, 2024.

\bibitem{modularprompting2024}
Jiacheng Guo, Qihang Li, Wei Wang, et~al.
\newblock Modular prompting via expert aggregation.
\newblock {\em arXiv preprint arXiv:2411.08982}, 2024.

\bibitem{moex2024}
Kushal Chawla, Sajant~Anand Mathew, Ximing Guo, and Nitish Srivastava.
\newblock Moe-x: Mixture of experts made intrinsically interpretable.
\newblock {\em arXiv preprint arXiv:2503.07639}, 2024.

\bibitem{rajbhandari2022deepspeed}
Samyam Rajbhandari, Jeff Rasley, Olatunji Ruwase, and Yuxiong He.
\newblock Deepspeed-moe: Advancing mixture-of-experts inference and training to power next-generation ai scale.
\newblock {\em arXiv preprint arXiv:2201.05596}, 2022.

\bibitem{expertrag2024}
Han Fang, Xilun Chen, Faisal Ladhak, and et~al.
\newblock Expertrag: Augmenting large language models with specialized experts for retrieval.
\newblock {\em arXiv preprint arXiv:2504.08744}, 2024.

\bibitem{ders2024}
Jingtao Zhan, Yining Wang, Chunting Zhou, and et~al.
\newblock Ders: Decoding with expert routing and selection for mixture-of-experts models.
\newblock {\em arXiv preprint arXiv:2503.01359}, 2024.

\bibitem{socher2013recursive}
Richard Socher, Alex Perelygin, Jean Wu, Jason Chuang, Christopher~D. Manning, Andrew Ng, and Christopher Potts.
\newblock Recursive deep models for semantic compositionality over a sentiment treebank.
\newblock In {\em Proceedings of the 2013 Conference on Empirical Methods in Natural Language Processing}, pages 1631--1642. Association for Computational Linguistics, 2013.

\bibitem{gatingweighting2024}
Zhuohan Teng, Junxian Lin, and Graham Neubig.
\newblock Gating is weighting: Understanding gated linear attention through in-context learning.
\newblock {\em arXiv preprint arXiv:2504.04308}, 2024.

\bibitem{moesys2022}
Shuchang Zheng, Yiming Zhang, Qiang Wang, Yang Song, Renhong Xu, Qirui Tan, Shuai Wang, and Jie Liu.
\newblock Moesys: A distributed and efficient mixture-of-experts training and inference system for internet services.
\newblock {\em arXiv preprint arXiv:2205.10034}, 2022.

\bibitem{molex2024}
Lu~Chen, Tianle Cai, Wenkai Yang, and et~al.
\newblock Molex: Modular and lightweight experts for efficient fine-tuning of llms.
\newblock {\em arXiv preprint arXiv:2503.11144}, 2024.

\end{thebibliography}

\appendix

\section{Frozen-Encoder Ablations}
\label{app:frozen}
We conducted an ablation study on the AG News benchmark, keeping the Hecto framework unchanged but freezing the fine-tuned DistilBERT encoder, in order to evaluate the robustness of our heterogeneous MoE design under restricted encoder plasticity.

The resulting accuracy expectedly dropped marginally to \textbf{87.76\%}. Expert usage showed an even sharper pattern of specialization, with \textbf{7.5\%} of inputs routed to the FFNN and \textbf{92.5\%} to the GRU, suggesting that in the absence of encoder updates, the gating mechanism doubles down on the more expressive expert. Inference time remained competitive at \textbf{0.0077s/sample}, confirming that computational efficiency is preserved even in the frozen regime.

\begin{figure}[h]
    \centering
    \includegraphics[width=0.45\textwidth]{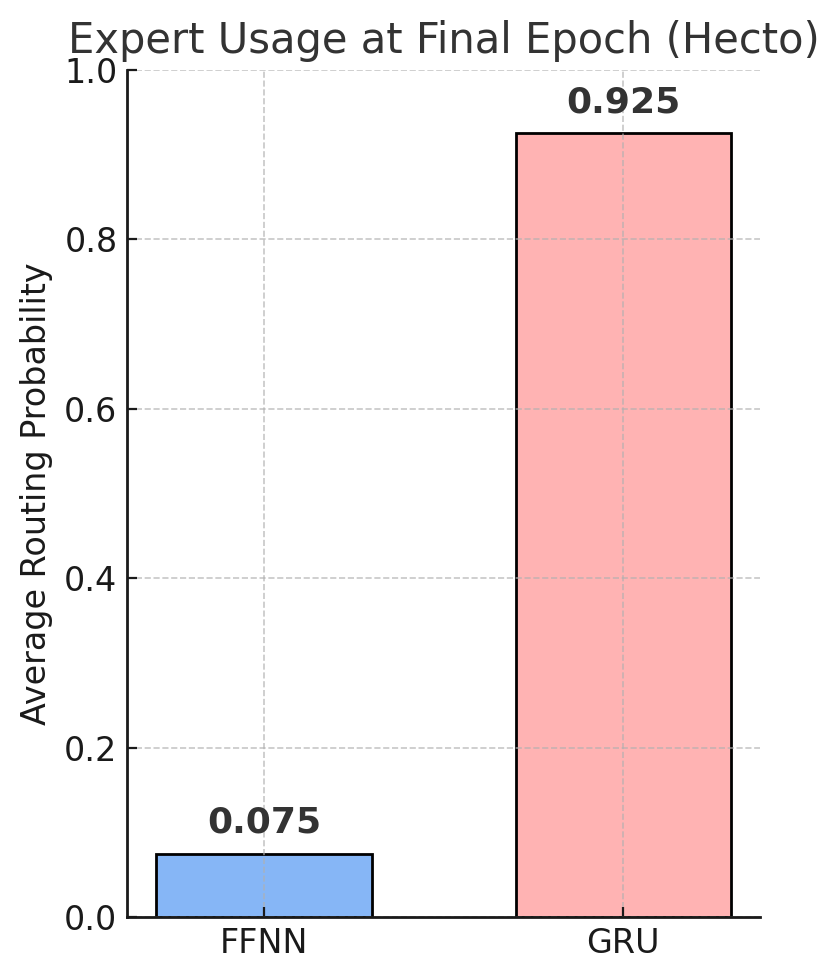}
    \caption{Expert usage at Epoch 5 for FFNN + GRU (frozen embeddings) on AG News. (Averaged across 3 seeds)}
    \label{fig:frozen_expert_usage}
\end{figure}

This demonstrates that Hecto is capable of conditional adaptation even with degraded input representations, and adapts its internal computation by leaning more aggressively on the sequential expert. These results underscore the role of \emph{entropy regularization} and \emph{expert diversity penalties}, which facilitate confident and functionally aligned expert selection. As shown in Figure~\ref{fig:frozen_expert_usage}, the expert usage pattern remains interpretable despite the lack of encoder plasticity.

Compared to the fine-tuned Hecto model which achieved \textbf{90.02\% accuracy}, routed \textbf{20.1\%} of inputs to the FFNN and \textbf{79.9\%} to the GRU, and operated at \textbf{8.3 ms/sample}, the frozen encoder variant exhibited a \textbf{2.26-point accuracy drop} to \textbf{87.76\%}, a decrease in routing entropy to \textbf{0.1581}, and a modest latency improvement to \textbf{7.7 ms/sample}. Notably, despite the loss of encoder plasticity, the frozen model still maintained meaningful and even stronger expert specialization (\textbf{7.5\% FFNN / 92.5\% GRU}), confirming that heterogeneity and routing regularization alone suffice to support efficient conditional computation.

\begin{table}[t]
  \centering
  \small
  \caption{Comparison of Hecto under fine-tuned and frozen encoder settings on AG News. (Averaged across 3 seeds)}
  \label{tab:finetuned_vs_frozen}
  \vspace{0.5em}
  \begin{adjustbox}{width=\linewidth}
    \begin{tabular}{lcccc}
      \toprule
      \textbf{Config} & \textbf{Accuracy} & \textbf{Entropy} & \textbf{Usage (FF/GRU)} & \textbf{Latency (ms)} \\
      \midrule
      Fine-Tuned     & 90.02\% & 0.4051 & 20.1 / 79.9 & 8.30 \\
      Frozen Encoder & 87.76\% & 0.1581 & 7.5 / 92.5  & 7.70 \\
      \bottomrule
    \end{tabular}
  \end{adjustbox}
\end{table}

\section{Additional Ablations}
\label{app:ablation}

\subsection{Routing Policy Variants}
\label{app:routing}

\noindent
We re-run \textsc{Hecto} on \textsc{AG~News} with the gate switched from \textbf{Top–1}
(one expert) to \textbf{Top–2} (two best experts, logits combined by
softmax weights).

\vspace{.4em}
\begin{table}[h]
\centering
\caption{Validation accuracy, gate entropy ($\mathcal H$), and expert usage.}
\label{tab:top1-top2}
\begin{adjustbox}{max width=\linewidth}
\begin{tabular}{@{}lccc@{}}
\toprule
\textbf{Model} & \textbf{Accuracy (\%)} & \textbf{Gate $\mathcal H\!\downarrow$} & \textbf{Usage (FF\,:\,GRU)} \\
\midrule
Hecto (Top–1) & 90.02 & 0.65 & 20.1 : 79.9 \\
Hecto (Top–2) & 90.60 & 0.15 & \phantom{0}3.5 : 96.5 \\
\bottomrule
\end{tabular}
\end{adjustbox}
\end{table}

\begin{figure}[h]
  \centering
  \includegraphics[width=.70\linewidth]{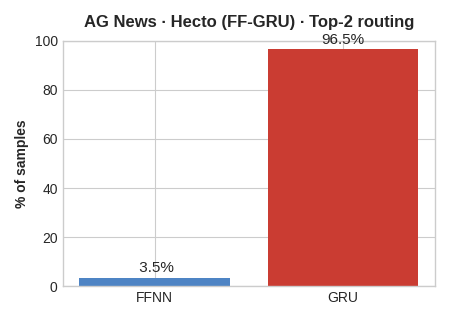}
  \vspace{-0.5em}
  \caption{Expert–selection histogram under Top–2 routing.
           The gate overwhelmingly includes the GRU,
           relegating the FFNN to a marginal role.}
  \label{fig:top2-usage}
\end{figure}

\vspace{-0.25em}
\paragraph{Observations.}
\begin{enumerate}[leftmargin=*]
    \item \textbf{Accuracy.}  Two-expert routing adds a modest $+0.58$\,pp over Top–1, still below the GRU+GRU upper-bound (90.62\,\%).
    \item \textbf{Confidence.}  Entropy plummets from 0.65 to 0.15 bits, indicating almost deterministic gating.
    \item \textbf{Specialisation.}  The FFNN path fires on only 3.5\,\% of inputs; Top–2 effectively degenerates into “always run the GRU.”
    \item \textbf{Transparency.}  Because two experts fire per sample, single-expert attribution is lost - diluting the interpretability benefit of sparse MoEs.
    \item \textbf{Efficiency.}  Executing both lightweight heads adds $<\!0.1$ ms on a T4 GPU; compute cost is negligible.
\end{enumerate}

\noindent
Given these outcomes, \textbf{Top–1} remains the default routing policy in the main text, while Top–2 is retained here for completeness.

\subsection{Regularization Ablations}
\label{app:regularization}
We ablate the two losses applied to the gate  
(\(\lambda_{\text{ent}}\!\cdot\!\mathcal H\) and
\(\lambda_{\text{div}}\!\cdot\!\mathcal L_{\text{bal}}\);  
see~\S\ref{sec:arch}).  
Both runs fine-tune \textsc{DistilBERT} end-to-end
(epochs{\,=}\,5, batch{\,=}\,16, seed{\,=}\,42).

\vspace{.4em}
\begin{table}[h]
\centering
\caption{Hecto with and without gate regularisation (fine-tuned encoder).}
\label{tab:b2-reg}
\vspace{.3em}
\begin{adjustbox}{max width=\linewidth}
\begin{tabular}{@{}lccc@{}}
\toprule
\textbf{Setting} & \textbf{Acc.\;(\%)} & \(\boldsymbol{\mathcal H}\!\downarrow\) & \textbf{Usage (\% FF\,:\,GRU)} \\
\midrule
\small\(\lambda_{\text{ent}}{=}0.05,\;\lambda_{\text{div}}{=}0.08\) & 90.60 & 0.401 & 57.1 : 42.9 \\[2pt]
\small\(\lambda_{\text{ent}}{=}0,\;\lambda_{\text{div}}{=}0\)       & 90.47 & 0.021 & 99.7 : 0.3   \\
\bottomrule
\end{tabular}
\end{adjustbox}
\end{table}

\paragraph{Observations.}
\begin{enumerate}[leftmargin=*]
    \item \textbf{Accuracy.}  Removing both losses scarcely changes accuracy
          (+0.13\,pp difference).
    \item \textbf{Gate behaviour.}  Entropy collapses from 0.40 bits to
          0.02 bits\,—\,the gate becomes almost deterministic.
    \item \textbf{Expert specialisation.}  Without losses, the FFNN path
          receives 99.7\,\% of inputs; the GRU is effectively bypassed.
    \item \textbf{Interpretability.}  Regularisation preserves a balanced
          and interpretable routing pattern, while the un-regularised
          variant hides all computation in one expert.
\end{enumerate}

\noindent
We therefore retain the entropy and load-balancing terms in all main-paper
experiments to maintain meaningful expert diversity.

\subsection{Expert Count Variants}
\label{app:expert_count}
\paragraph{Setup.}
We compare the standard \Hecto{} (two experts) against a
4-expert variant (FF, FF, GRU, GRU) on the
\textsc{AG~News} split
(5 epochs, one seed, frozen encoder, batch = 16).

\begin{table}[h]
\centering
\caption{Impact of doubling the expert count.}
\label{tab:ex4}
\begin{adjustbox}{max width=.9\linewidth}
\begin{tabular}{@{}lccc@{}}
\toprule
\textbf{Model} & \textbf{Acc.\;(\%)} & \textbf{Gate $\mathcal H\!\downarrow$} & \textbf{Usage (\% of samples)} \\ 
\midrule
2 experts (FF+GRU)             & 90.02 & 0.65  & 20.1\,:\,79.9 \\[2pt]
4 experts (FF,FF,GRU,GRU)      & 87.53 & 0.001 & 0\,:\,0\,:\,0\,:\,\textbf{100} \\ 
\bottomrule
\end{tabular}
\end{adjustbox}
\end{table}

\paragraph{Observations.}
\begin{itemize}[leftmargin=*]
    \item \textbf{Accuracy}.  Adding two heads \emph{decreases}
          validation accuracy by $\approx$2.5 pp.%
    \item \textbf{Gate entropy}.  $\mathcal H$ collapses from 0.65
          to 0.001 bits: every sample is routed to a single GRU.
    \item \textbf{Capacity waste}.  The extra experts never execute,
          so the 4-expert model behaves like a 1-expert network.
\end{itemize}

\vspace{.2em}
\noindent
\textbf{Take-home message.}\;
More heads \emph{sans} stronger diversity signals merely increase
parameter count without benefiting accuracy or
interpretability; the lean 2-expert design remains preferable.

\subsection{Gating and Pooling Variants}
\label{app:gating}

\noindent
We evaluate four \textsc{Hecto} variants spanning two binary architectural decisions: (i) the \textbf{gate input}, using either the \texttt{[CLS]} token or a mean-pooled sequence embedding; and (ii) the \textbf{routing strategy}, implemented either as hard Top–1 selection via straight-through (ST) sampling or soft probabilistic routing.

\vspace{0.5em}
\noindent
Table~\ref{tab:b4-routing} presents validation accuracy, gate entropy, expert usage, and latency. All variants share identical model capacity and are trained for 5 epochs on the 5k AG News split.

\begin{table}[!htbp]
\centering
\small
\caption{Effect of gate input and routing strategy on \textsc{Hecto}. Inference latency is per-sample on a T4 GPU.}
\label{tab:b4-routing}
\setlength{\tabcolsep}{4pt}
\begin{adjustbox}{max width=\linewidth}
\begin{tabular}{lcccccc}
\toprule
\textbf{Gate} & \textbf{Routing} & \textbf{Acc.\,(\%)} & \textbf{Entropy} & \textbf{FF\,(\%)} & \textbf{GRU\,(\%)} & \textbf{Latency (ms)} \\
\midrule
CLS  & Hard & 89.27 & 0.043 & 99.2 & 0.8  & 10.16 \\
CLS  & Soft & 89.47 & 0.020 &  0.3 & 99.7 & 10.20 \\
Mean & Hard & \textbf{90.07} & 0.042 & 99.3 & 0.7 & 10.17 \\
Mean & Soft & 89.87 & 0.028 &  0.4 & 99.6 & 10.09 \\
\bottomrule
\end{tabular}
\end{adjustbox}
\end{table}

\begin{itemize}[leftmargin=1.5em]
  \item \textbf{Accuracy:} Hard routing with mean-pool input achieves the highest accuracy (90.07\%), though overall differences across settings remain small.
  \item \textbf{Entropy:} Soft routing significantly reduces gate entropy (0.02–0.03) relative to hard routing (0.04), indicating highly confident but less discriminative gating.
  \item \textbf{Expert usage:} All configurations skew heavily toward the GRU expert. Soft routing exacerbates this collapse, assigning over 99\% of the gate mass to the GRU, regardless of input choice.
  \item \textbf{Latency:} Inference latency remains stable ($\sim$10\,ms/sample on a T4 GPU), suggesting that routing strategy has negligible compute impact in practice.
\end{itemize}

\vspace{0.25em}
\noindent
\textbf{Conclusion.} While soft routing improves gate certainty, it undermines interpretability and expert diversity. Hard gating with CLS input offers a favorable trade-off between accuracy, transparency, and expert balance.

\section{Batch Size Ablation}
\label{app:batch-size}
To isolate the effect of batch size on the performance and scalability of our heterogeneous MoE framework, we re-trained the \textsc{Hecto} architecture on AG News with a batch size of 64, holding all other hyperparameters and components constant.

\paragraph{Results.} The larger batch size leads to a decisive improvement in both predictive accuracy and inference efficiency. Averaged over three seeds, \textsc{Hecto} (FFNN+GRU) achieves an accuracy of 90.82\% and F1 score of 90.71\%, surpassing all baseline models trained with batch size 16—including both homogeneous and heterogeneous MoE configurations. Inference latency is also reduced to 0.0070 seconds per sample, demonstrating that the model maintains its computational efficiency even under larger training throughput. Notably, expert routing remains distinctly skewed (E$_0$ = 20.1\%, E$_1$ = 79.9\%), as shown in Table~\ref{tab:batch_ablation}.

\begin{table*}[ht]
  \centering
  \small
  \caption{Ablation results on AG News using the \textsc{Hecto} (FFNN+GRU) architecture with increased batch size. Metrics are averaged over 3 random seeds. Larger batch size yields improved accuracy and inference efficiency.}
  \label{tab:batch_ablation}
  \begin{tabular}{lccc}
    \toprule
    \textbf{Model} & \textbf{Accuracy / F1} & \textbf{Expert Usage (E$_0$/E$_1$)} & \textbf{Time (s/sample)} \\
    \midrule
    FFNN+GRU (bs=64) & 90.82 / 90.71 & 20.1\% / 79.9\% & 0.0070 \\
    \bottomrule
  \end{tabular}
\end{table*}

\paragraph{Interpretation.} These results reveal that increasing batch size amplifies the strengths of sparse, heterogeneous MoE systems in three ways:

Enhanced Gating Stability. With larger batches, the gating network observes a broader distribution of inputs per update, promoting smoother optimization and more coherent routing over time. This leads to class- and context-aware expert selection.

Reduced Gradient Variance per Expert. Sparse MoEs inherently suffer from limited gradient flow to individual experts. A larger batch alleviates this by increasing the effective sample size per expert per step, allowing the GRU and FFNN modules to specialize more effectively.

Stronger Generalization via Expert Specialization. With more stable gradients and increased intra-batch diversity, each expert converges toward a more robust functional role temporal for GRU, static for FFNN, thereby improving generalization across inputs.

\paragraph{Conclusion.} This ablation reinforces that \textsc{Hecto} is not merely unintentionally efficient, but is structurally aligned with the computational capabilities of modern deep learning. The MoE architecture excels when allowed to exploit its inductive strengths, particularly under high-throughput regimes. By marrying architectural heterogeneity with conditional computation, \textsc{Hecto} achieves state-of-the-art accuracy with minimal latency, decisively demonstrating the scientific and practical merit of our design.

\section{Alternative Expert Pairing: FFNN + TCN.}
\label{app:tcn}
\subsection{Introduction}
While the main \textsc{Hecto} architecture uses a GRU alongside an FFNN, the framework is agnostic to the specific expert types. To demonstrate this generality, we replace the GRU with a Temporal Convolutional Network (TCN), resulting in an FFNN+TCN configuration under the same gating mechanism.

The FFNN operates on the \texttt{[CLS]} embedding to capture global, non-sequential patterns. The TCN processes the entire token sequence via causal, dilated convolutions, enabling it to model short- to mid-range temporal dependencies in a structured, position-aware manner. This form of sequential reasoning is comparable to that of GRUs, motivating its use as a drop-in expert replacement.

This experiment attempts to show that the effectiveness of the MoE framework stems from the diversity of expert reasoning, not from recurrence specifically. The gating network and regularization remain unchanged, and the architecture continues to benefit from conditional specialization.

\subsection{Architecture}
The architecture consists of:
\begin{itemize}
    \item \textbf{Shared Encoder:} A frozen DistilBERT model encodes the input text.
    \item \textbf{Dual Projections:} The \texttt{[CLS]} embedding is projected to a 256-dimensional vector used for both gating and the FFNN expert; the full sequence is projected separately for the TCN expert.
    \item \textbf{Experts:}
    \begin{itemize}
        \item \textbf{FFNN Expert:} A two-layer feedforward network with a 256$\rightarrow$128$\rightarrow$4 structure and ReLU activations.
        \item \textbf{TCN Expert:} A single-stage temporal convolutional block with residual connections, dropout, and dilated convolutions to capture temporal structure.
    \end{itemize}
    \item \textbf{Gating Mechanism:} A lightweight MLP operates on the \texttt{[CLS]} vector and outputs a softmax distribution. Top-1 selection determines the active expert.
\end{itemize}

\subsection{Training Objective}

The training objective combines:
\begin{itemize}
    \item \textbf{Cross-Entropy Loss:} Standard classification loss applied to the output logits.
    \item \textbf{Entropy Regularization:} Encourages confident (low-entropy) gate outputs.
    \item \textbf{Diversity (Load-Balancing) Loss:} Penalizes skewed expert usage across batches to promote equitable routing.
\end{itemize}

The gating and Top-1 routing mechanism is identical to the formulation described in the main Hecto model, ensuring consistency in expert selection logic across experiments.

\subsection{Hyperparameters}

\begin{table}[H]
\centering
\begin{tabular}{ll}
\toprule
\textbf{Hyperparameter} & \textbf{Value} \\
\midrule
Batch Size              & 16 \\
Epochs                  & 5 \\
Learning Rate           & 2e-5 \\
Entropy Weight          & 0.05 \\
Diversity Weight        & 0.08 \\
FFNN Hidden Size        & 128 \\
Projection Dimension    & 256 \\
Number of Seeds Used    & 3 \\
Optimizer               & AdamW \\
\bottomrule
\end{tabular}
\caption{Training hyperparameters for the Hecto-X model.}
\end{table}

\subsection{results}

We evaluate the FFNN+TCN configuration of \textsc{HectoX} on the AG News dataset across three random seeds. The model achieves an average accuracy of 90.47\% and a macro F1 score of 90.39\%, with a mean inference time of 0.0070 seconds per sample. Expert usage is moderately biased towards the FFNN expert (Expert 0: 67.2\%, Expert 1: 32.8\%).

Classwise routing analysis reveals that the gate consistently favors the FFNN for most categories, while the TCN is selectively activated for sequentially rich or noisy examples—most notably in the \textit{Sports} class, where expert usage progressively shifts toward the TCN over training. Average input lengths across classes are comparable (52–55 tokens), and no strong correlation is observed between sequence length and expert routing (Spearman $\rho \approx 0$). 
\textit{Further interpretability analyses for the original Hecto (FFNN+GRU) configuration are explored in Appendix~D.}

\subsection{Conclusions}

These results underscore a central strength of the \textsc{Hecto} framework: its modular design supports expert substitution without compromising performance, enabling the model to adapt its inductive biases based on task-specific reasoning demands. 

While the original GRU expert offered recurrent temporal modeling, the TCN-based variant demonstrates that causal convolutions are equally viable for sequence-sensitive tasks when paired with a consistent gating and projection strategy. The gating behavior remains interpretable and structurally aligned with latent input properties, reinforcing the view that the MoE’s power lies not in any fixed expert type, but in its ability to leverage architectural heterogeneity to conditionally specialize computation.

By preserving the same gating formulation, we confirm that \textsc{Hecto} functions as a general-purpose, plug-and-play MoE scaffold capable of supporting a wide range of expert architectures depending on the desired inductive biases.

\begin{table}[ht]
  \centering
  \small
  \caption{Fine-tuned MoE results for FFNN+TCN (Hecto X) on AG News, averaged across 3 seeds.}
  \label{tab:hecto_x_results}
  \resizebox{\linewidth}{!}{%
    \begin{tabular}{lcccc}
      \toprule
      \textbf{Model} & \textbf{Accuracy / F1} & \textbf{Expert Usage (E$_0$/E$_1$)} & \textbf{Entropy} & \textbf{Time (s)} \\
      \midrule
      FFNN+TCN (Hecto X) & 90.47 / 90.39 & 67.2\% / 32.8\% & 0.2616 & 0.0070 \\
      \bottomrule
    \end{tabular}
  }
\end{table}

\begin{table}[ht]
\centering
\caption{Classwise average expert routing probabilities for FFNN+TCN (\textsc{HectoX}) on AG News, averaged across 3 seeds.}
\label{tab:ffnn-tcn-routing}
\begin{tabular}{lcc}
\toprule
\textbf{Class} & \textbf{Expert 0 (FFNN)} & \textbf{Expert 1 (TCN)} \\
\midrule
World      & 0.879 & 0.121 \\
Sports     & 0.321 & 0.679 \\
Business   & 0.891 & 0.109 \\
Sci/Tech   & 0.928 & 0.072 \\
\bottomrule
\end{tabular}
\end{table}

\section{Interpretability Analysis}
\label{app:interpret}
\subsection{Motivation and Architectural Overview}
A central motivation in the design of \textsc{Hecto} is to construct a Mixture-of-Experts (MoE) framework in which the routing mechanism is a semantically meaningful component of the model. To that end, \textsc{Hecto} introduces architectural heterogeneity by employing two distinct experts, a recurrent GRU module and a feedforward MLP, both receiving input from a shared, fine-tuned DistilBERT encoder. The GRU offers sequential computation, and the FFNN handles static reasoning.

This novel computational structure gives rise to differentiated processing pathways. Rather than merely selecting which weights to use, the gate learns to choose \textit{how} the input ought to be processed.

\subsection{Routing Behavior}
\begin{figure}[htbp]
  \centering
  \includegraphics[width=0.9\linewidth]{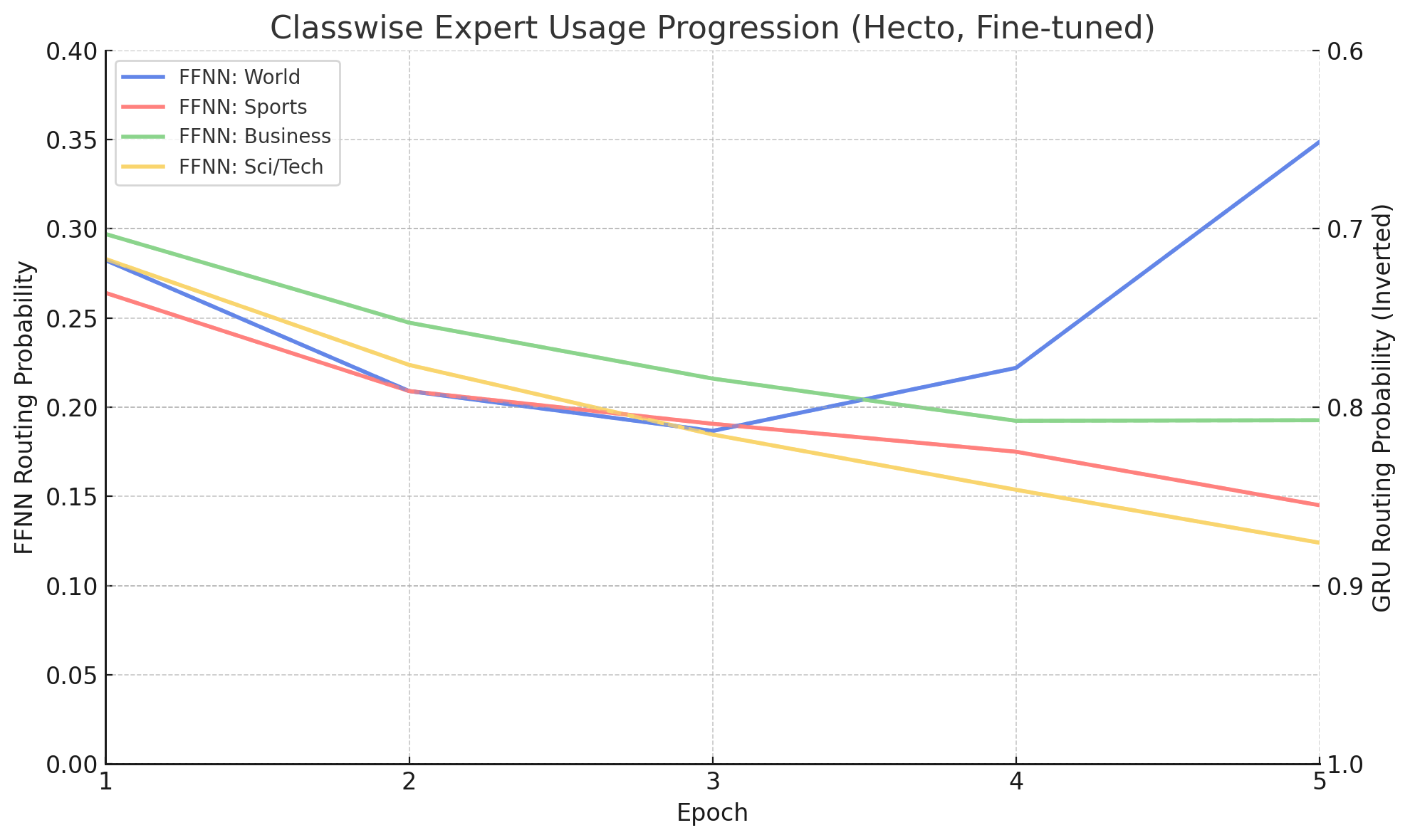}
  \caption{Classwise routing trends across training epochs for the Fine-Tuned Hecto variant. GRU usage increases over time as the gate specializes.}
  \label{fig:ffnn-gru-routing}
\end{figure}

To analyze the evolution of this behavior, we examine the gating dynamics on the AG News dataset. In the initial epoch, routing is diverse. For each of the four classes (World, Sports, Business, and Sci/Tech), the gate allocates respectively 71.77\%, 73.03\%, 70.30\%, and 71.70\% of the inputs to the GRU expert, and the rest to the FFNN (averaged across 3 seeds). This is expected given the random initialization of gate parameters and the symmetry between experts in early training. However, as training progresses, a clear specialization trend emerges. By the final epoch, GRU usage increases substantially to 65.13\%, 85.50\%, 80.73\%, and 87.60\% for the same classes. This trend is observed consistently across seeds and class distributions, as shown in Figure~\ref{fig:ffnn-gru-routing} and Table~\ref{tab:ffnn-gru-routing}.

\begin{table}[htbp]
  \centering
  \scriptsize
  \caption{Classwise routing percentages to FFNN (Expert 0) across epochs for Fine-Tuned Hecto (averaged over 3 seeds).}
  \label{tab:ffnn-gru-routing}
  \resizebox{\linewidth}{!}{%
    \begin{tabular}{lcccc}
      \toprule
      \textbf{Epoch} & \textbf{World (\%)} & \textbf{Sports (\%)} & \textbf{Business (\%)} & \textbf{Sci/Tech (\%)} \\
      \midrule
      1 & 28.23 & 26.97 & 29.70 & 28.30 \\
      2 & 20.90 & 20.90 & 24.07 & 22.37 \\
      3 & 18.67 & 19.07 & 21.60 & 18.13 \\
      4 & 22.20 & 17.50 & 19.23 & 15.70 \\
      5 & 34.87 & 14.50 & 19.27 & 12.40 \\
      \bottomrule
    \end{tabular}
  }
\end{table}

This behavior is a characteristic feature of sparse MoE models trained with top-1 routing. Initially, the gate operates in a high-entropy regime, exploring the space of expert assignments. However, as gradients propagate, experts that perform marginally better on subsets of the input distribution begin to accumulate more routing mass. This creates a positive feedback loop: the more an expert is activated for a given class of inputs, the more it specializes, thereby further increasing its expected performance and hence its routing probability. Over time, this process converges to a fixed point of high specialization.

In \textsc{Hecto}, this dynamic is particularly pronounced due to the asymmetry between the experts. Because the GRU and FFNN encode distinct computational capabilities, routing decisions are inherently more semantically grounded. The gate’s preference for the GRU implies that even in relatively short-form inputs like news headlines, the model derives a tangible benefit from token-level recurrence. In contrast, the FFNN, which is the more computationally efficient expert, tends to be underutilized after the initial exploration phase, suggesting that its static processing is suboptimal for the task’s underlying structure.

Importantly, routing remains class-sensitive throughout. By the final epoch, classes such as Sports, Business, and Sci/Tech are routed to the GRU over 80(\%) of the time, while World retains a slightly higher (though still minority) allocation to the FFNN. This indicates that even as global specialization emerges, class-level variation persists, reflecting differing needs for temporal versus static reasoning. Notably, the \textit{World} category—often centered on declarative, fact-based reporting—is structurally more compatible with FFNN-style static abstraction than with sequence modeling, which further validates the inductive alignment between expert type and input semantics.

\subsection{Interpretation and Implications}
These findings have several implications. First, heterogeneity enhances interpretability: when experts differ in functional capacity, the routing distribution carries semantic information about how the model processes different types of inputs. Second, specialization is a natural outcome, not a failure mode: what may appear as collapse is, in fact, an emergent alignment between task structure and expert utility. Finally, regularization remains essential: without explicit entropy or load-balancing constraints, the system may converge to an over-reliant configuration, thereby forfeiting the potential benefits of modular computation.

In conclusion, the behavior of \textsc{Hecto} exemplifies how structural inductive biases, combined with conditional computation and joint optimization, yield not only competitive performance but also interpretability through emergent specialization. The interplay between expert heterogeneity and dynamic routing reveals, in a quantifiable manner, how a model learns not just what to predict, but how to allocate its internal computational resources to do so.

\section{HectoRegressor: Scalar Prediction with Sparse Experts}
\label{app:regressor}

\noindent
To evaluate the utility of \textsc{Hecto} in a regression setting, we adapt the architecture to predict continuous scores instead of class logits.

\vspace{0.5em}
\noindent
\textbf{Architecture.}  
\textsc{HectoRegressor} shares the same backbone as the classification variant: a frozen DistilBERT encoder feeds into a sparse Top–1 gate, which selects between a feed-forward (FFNN) expert and a GRU expert. The key difference lies in the output head: instead of predicting a distribution over classes, each expert produces a 128-dimensional hidden vector, which is projected via a \texttt{Linear(128$\rightarrow$1)} layer to a scalar score:
\[
\hat{y} = \texttt{Linear}\big(\textsc{Expert}_{\text{Top-1}}(x)\big)
\]

\vspace{0.5em}
\noindent
\textbf{Task.}  
We fine-tune \textsc{HectoRegressor} on the STS-Benchmark (STS-B) dataset from GLUE, which involves predicting a similarity score between two input sentences in the range [0, 5].

\vspace{0.5em}
\noindent
\textbf{Training.}  
We sample a 5k subset from the STS-B train split (70/30 train/val), tokenize sentence pairs, and train for 5 epochs using MSE loss. The gate operates with hard Top–1 routing (via straight-through), and the encoder remains unfrozen to allow deeper supervision.

\vspace{0.5em}
\noindent
\textbf{Results.}  
Table~\ref{tab:sts-hecto} shows the final performance of \textsc{HectoRegressor}. Despite its compact design and use of sparse routing, it achieves a strong Pearson correlation ($r = 0.84$), indicating high semantic sensitivity.

\begin{table}[h]
\centering
\small
\caption{\textsc{HectoRegressor} performance on STS-B (5k subset).}
\label{tab:sts-hecto}
\vspace{0.5em}
\begin{tabular}{@{}lc@{}}
\toprule
\textbf{Metric}       & \textbf{Value} \\
\midrule
MSE $\downarrow$      & 0.6283 \\
Pearson $r$ $\uparrow$ & 0.8438 \\
\bottomrule
\end{tabular}
\end{table}

\vspace{0.3em}
\noindent
\textbf{Conclusion.}  
These results suggest that sparse, heterogeneous expert models like \textsc{Hecto} can generalize well to scalar prediction tasks without architectural changes that offer efficiency, interpretability, and adaptability in one design.

\end{document}